\documentclass[10pt,twocolumn,letterpaper]{article}
\usepackage[table,xcdraw]{xcolor}
\usepackage{float}
\usepackage{cvpr}              

\newcommand{\red}[1]{{\color{red}#1}}

\definecolor{cvprblue}{rgb}{0.21,0.49,0.74}
\usepackage[pagebackref,breaklinks,colorlinks,linkcolor=red,citecolor=cvprblue]{hyperref}
\usepackage{amssymb}
\usepackage{mathrsfs}
\usepackage{amsthm,amsmath,amssymb}
\usepackage{tabularx}
\usepackage{cuted}
\usepackage{cancel}

\def\paperID{735} 
\def\confName{CVPR}
\def\confYear{2025}

\title{One-Way Ticket\includegraphics[height=20pt, trim=-0.5cm 2.0cm -0.5cm 0cm]{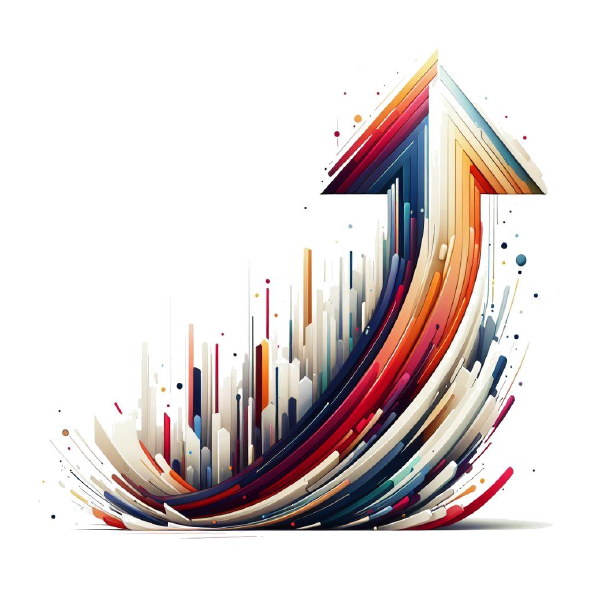}: Time-Independent Unified Encoder for Distilling \\ Text-to-Image Diffusion Models}

\usepackage{amsmath}
\usepackage{amssymb}
\usepackage{mathtools}
\usepackage{multirow}
\usepackage{makecell}
\usepackage[capitalize]{cleveref}
\crefname{section}{Sec.}{Secs.}
\Crefname{section}{Section}{Sections}
\crefname{table}{Tab.}{Tabs.}
\Crefname{table}{Table}{Tables}
\usepackage{graphicx}
\usepackage{booktabs}
\usepackage{wrapfig}
\usepackage{enumitem}
\usepackage{diagbox}
\usepackage{xspace}

\usepackage{pifont}
\usepackage{multirow}
\usepackage{wrapfig,lipsum}
\usepackage{diagbox}
\usepackage{caption}

\usepackage[accsupp]{axessibility}  

\newcommand{\cmarkg}{\textcolor{green}{\ding{51}}\xspace}%
\newcommand{\xmarkg}{\textcolor{red}{\ding{55}}\xspace}%

\def\ourmethod{{\textit{TiUE}}\xspace}

\newcommand{\minisection}[1]{\vspace{0.02in}\noindent{\bf #1}}

\newcommand{\blue}[1]{{\color{blue}#1}}
\newcommand{\green}[1]{{\color{green}#1}}

\author{
Senmao Li$^{1}$  \quad Lei Wang$^{1}$ \quad Kai Wang$^{2}$ \quad Tao Liu$^{1}$  \quad Jiehang Xie$^{3}$ \quad Joost van de Weijer$^{2}$ \\ \quad Fahad Shahbaz Khan$^{4,5}$ \quad Shiqi Yang$^{6}$ \quad Yaxing Wang$^{1,7\thanks{The corresponding author.}}$ \quad Jian Yang$^{1}$\\
	$^1${VCIP, CS, Nankai University} \; $^2${Computer Vision Center, Universitat Aut\`onoma de Barcelona}\\$^3${School of Big Data and Computer Science, Guizhou Normal University}\\$^4${Mohamed bin Zayed University of AI}\; $^5${Linkoping University}\; $^6${SB Intuitions, SoftBank}\\
    $^7${Nankai International Advanced Research Institute (Shenzhen Futian), Nankai University}\\
	{\tt\small  \{senmaonk,scitop1998,ltolcy0,shiqi.yang147.jp\}@gmail.com jiehangxie@gznu.edu.cn}\\ {\tt\small {\{kwang,joost\}@cvc.uab.es} {fahad.khan@liu.se} {\{yaxing,csjyang\}@nankai.edu.cn}}
\vspace{-5mm}
}

\begin{document}
\maketitle

\begin{figure*}[t]
    \centering
    \includegraphics[width=0.999\linewidth, trim=0 0 0 0, clip]{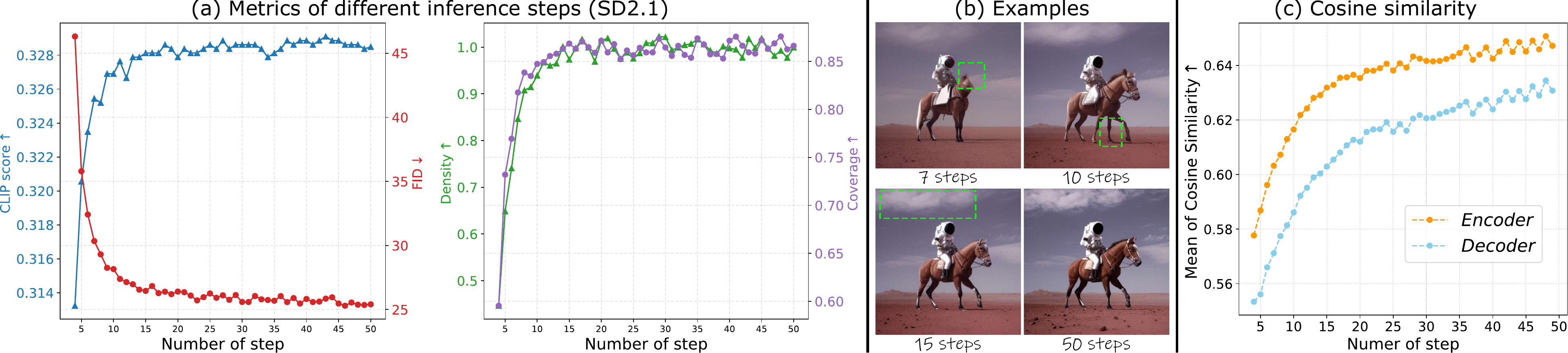}
    \vspace{-7mm}
    \caption{The correlation between image generation quality ({Fig.a-b}) and encoder feature similarity ({Fig.c}). Above a certain threshold of steps, such as 15 steps in SD2.1, the model maintains image generation quality (Fig.a-b) while the features show high similarity (Fig.c). Below this threshold, feature similarity deteriorates along with worse generation quality, accompanied by a degradation in image generation quality. Furthermore, the encoder features consistently exhibit higher similarity than the decoder across all sampling steps (Fig.c).}
    \vspace{-6mm}
      \label{fig:feat_similarity}
\end{figure*}

\begin{abstract}
Text-to-Image (T2I) diffusion models have made remarkable advancements in generative modeling; however, 
they face a trade-off between inference speed and image quality, posing challenges for efficient deployment.
Existing distilled T2I models can generate high-fidelity images with fewer sampling steps, but often struggle with diversity and quality, especially in one-step models. 
From our analysis, we observe redundant computations in the UNet encoders. 
Our findings suggest that, for T2I diffusion models, decoders are more adept at capturing richer and more explicit semantic information, while encoders can be effectively shared across decoders from diverse \textit{time steps}.
Based on these observations, we introduce the first Time-independent Unified Encoder (\ourmethod) for the student model UNet architecture, which is a loop-free image generation approach for distilling T2I diffusion models. Using a one-pass scheme, \ourmethod shares encoder features across multiple decoder time steps, enabling parallel sampling and significantly reducing inference time complexity. In addition, we incorporate a KL divergence term to regularize noise prediction, which enhances the perceptual realism and diversity of the generated images. Experimental results demonstrate that \ourmethod outperforms state-of-the-art methods, including LCM, SD-Turbo, and SwiftBrushv2, producing more diverse and realistic results while maintaining the computational efficiency.
\href{https://github.com/sen-mao/Loopfree}{https://github.com/sen-mao/Loopfree}
\end{abstract}

\vspace{-6mm}
\section{Introduction}
\label{sec:introduction}
Recently, diffusion models~\cite{song2021ddim,ho2020ddpm,dhariwal2021diffusionbeatgans} have achieved remarkable breakthroughs, representing a significant advance in the field of generative models. 
It is widely applied in diverse applications,
including image generation~\cite{ho2020ddpm,song2020score,rombach2022high,IF,liu2025onepromptonestory,hu2024tome}, video synthesis~\cite{text2video-zero,luo2023videofusion,wu2023tune,tokenflow2023}, image editing~\cite{hertz2022prompt,li2023stylediffusion,zhang2023adding,mou2023t2i,kai2023DPL}, T2I personalization~\cite{ruiz2023dreambooth,kumari2023multi,butt2025colorpeel,gu2023photoswap,wang2024mcti}, etc.
Despite their considerable success, diffusion models are not exempt from limitations. 
One of the main limitations of diffusion models is the slow inference speed.
This limitation affects the scalability and application of diffusion models in real-time environments.

\begin{figure*}[t]
\centering
\includegraphics[width=0.999\linewidth]{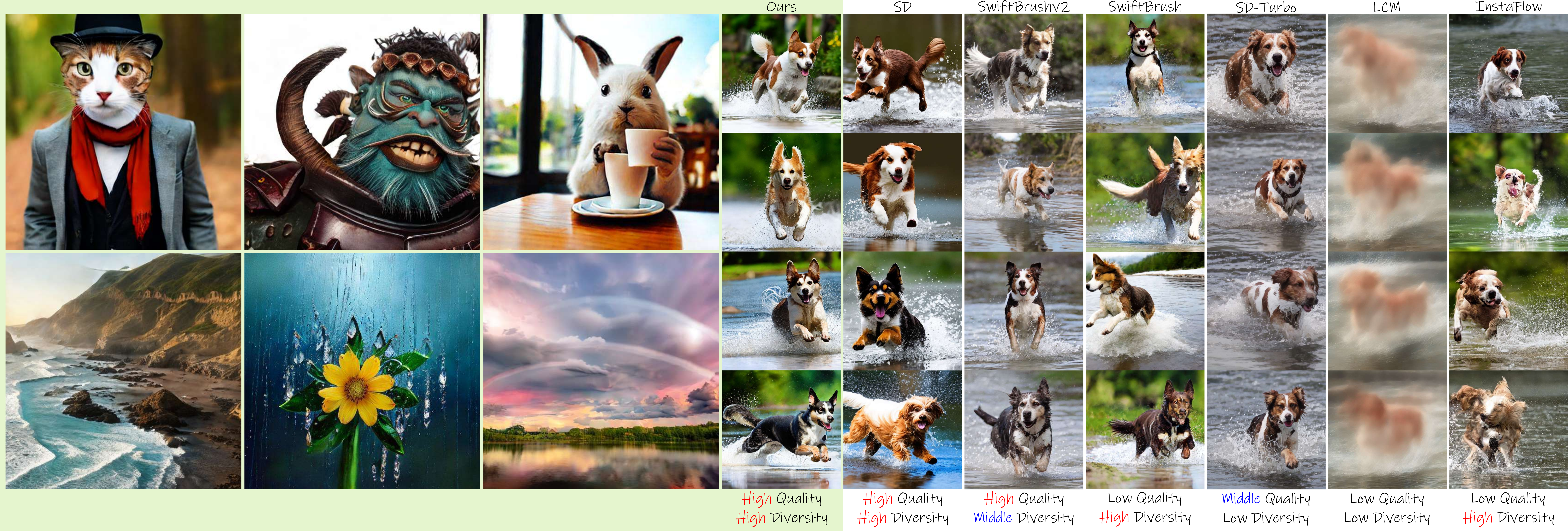}
\vspace{-6.5mm}
\caption{Generated high-fidelity $512^2$ images by the one-step model distilled with our method \ourmethod.
Compared with the baselines, our method produces higher image quality and more diverse results. Here the \textit{diversity} means the degree of variation of different output images with same semantic information when given the same prompt and different seed values. 
SwiftBrushv2~\cite{dao2024swiftbrushv2}, which is initialized with SD-Turbo~\cite{sauer2023adversarial}, achieves T2I generation with limited diversity as the SD-Turbo model.
}
\vspace{-6mm}
\label{sampling}
\end{figure*}

To address the limitations of T2I diffusion models during inference, current approaches achieve acceleration through either training-free techniques~\cite{su2024f3_pruning,shang2023post_quant,bolya2023token_merging} or training-based methods~\cite{salimans2022progressive,luo2023lcm_lora,sauer2023adversarial}. Training-free methods primarily focus on improving sampling solvers~\cite{song2021ddim,lu2022dpm} or using caching mechanisms to speed up individual sampling steps~\cite{ma2023deepcache,li2024faster}. Although these methods can reduce the number of sampling steps, even strong solvers still require more than 10 steps~\cite{lu2022dpm++}, limiting their efficiency due to their training-free nature.
On the other hand, training-based approaches distill a student generator~\cite{song2023consistency,luo2023latent,zheng2024TCD,sauer2023adversarial} from pretrained T2I diffusion models. Methods like LCM~\cite{luo2023latent}, SD-Turbo~\cite{sauer2023adversarial} and SwiftBrushv2~\cite{dao2024swiftbrushv2} can generate high-fidelity images with only a few sampling steps (e.g., 4 steps for LCM and 1 step for SwiftBrushv2). However, these approaches still struggle to produce high-quality and diverse images in the \textit{single} step generation. In this case, LCM produces low-quality generations, while SD-Turbo and SwiftBrushv2 exhibit reduced diversity, as observed in ~\cref{sampling},~\cref{fig:qualitative_comparison} and~\cref{fig:diversity}. 

Existing methods overlook redundant computations inherent in the UNet, an area that has been more thoroughly explored in training-free acceleration techniques~\cite{li2024faster,ma2023deepcache}. 
For example, Faster Diffusion~\cite{li2024faster} leverages feature similarity across adjacent inference steps by sharing the encoder across multiple decoder steps. We also observe that the correlation between image generation quality and encoder feature similarity remains effective up to a certain threshold as 15 steps for the SD model~\cite{rombach2022high}\footnote{This observation is based on statistical experiments on 100 generated images.}. 
Above this threshold, high-quality images are associated with high encoder feature similarity. Below this threshold, the image generation quality deteriorates ({\cref{fig:feat_similarity}(a-b)}) while the encoder feature exhibits variations  ({\cref{fig:feat_similarity}(c)}).
Additionally, the decoder feature always shows much higher variations than encoder (\cref{fig:feat_similarity}-(c)).

\begin{figure}[t]
    \centering
    \includegraphics[width=0.999\linewidth, trim=0 0 0 10, clip]{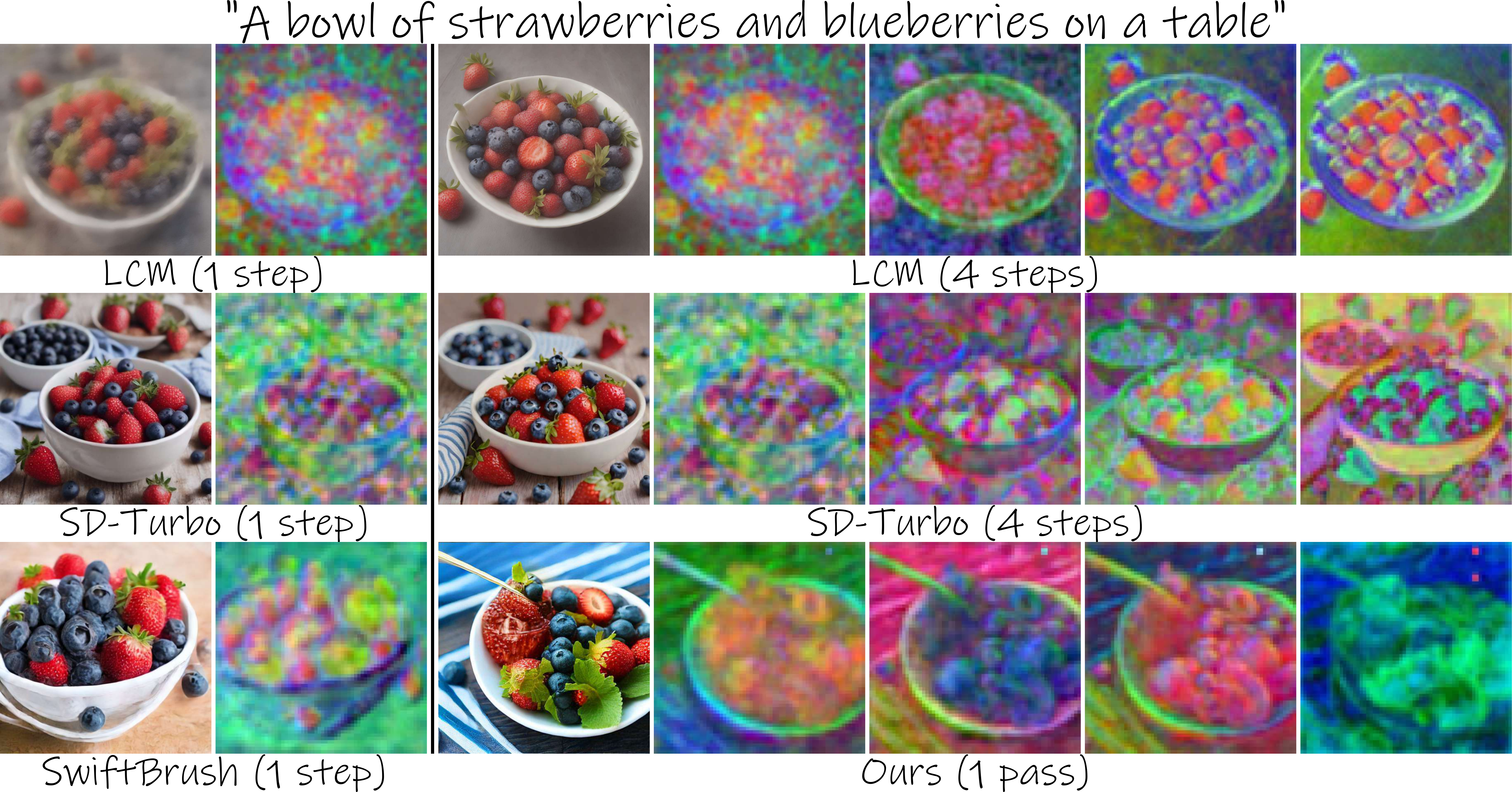}
    \vspace{-7mm}
    \caption{The generated images and the corresponding UNet decoder features using different time steps. The images generated in one step mainly focus on structural information, with a minimal amount of semantic information present in the feature layers (Left). By comparison, the decoder across multiple time-steps contain richer and more explicit semantic information in the feature level (Right), leading to better image generation.}
    \vspace{-6mm}
      \label{fig:features}
\end{figure}

Furthermore, we argue that this phenomenon also exists in distilled few-step diffusion models.
As observed in~\cref{fig:features}, the decoders of the 4-step generative models retrieve richer semantic information across diverse time steps. We hypothesize that this is the reason 4-step diffusion models generally achieve better generation performance compared to their 1-step counterparts. This hypothesis is also highly aligned with the finding in PnP~\cite{tumanyan2023plug} and DIFT~\cite{tang2023emergent}.

\begin{figure}[t]
\vspace{2mm}
\begin{center}
\includegraphics[width=0.999\linewidth]{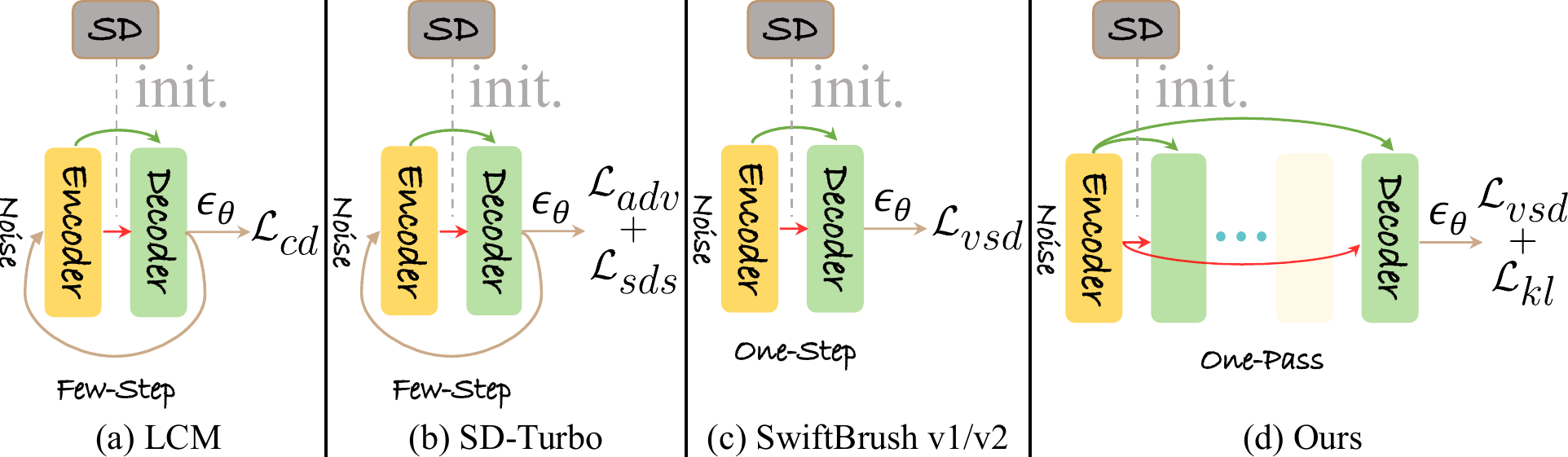}
\vspace{-6mm}
\caption{
Previous methods use the time-dependent UNet encoder-decoder design and differ from the distillation losses.
(a) LCM~\cite{luo2023latent} updates with consistency distillation loss $\mathcal{L}_{cd}$. 
(b) SD-Turbo~\cite{sauer2023adversarial} is trained with adversarial loss $\mathcal{L}_{adv}$ and score distillation loss $\mathcal{L}_{sds}$.
(c) SwiftBrushv1/v2~\cite{nguyen2023swiftbrush,dao2024swiftbrushv2} distills with variational score distillation $\mathcal{L}_{vsd}$ for one-step model. 
(d) Our method \ourmethod first proposes the Time-independent Unified Encoder (\ourmethod) design to achieve the one-pass loop-free generation. 
\ourmethod distills the T2I model with variational score distillation $\mathcal{L}_{vsd}$ and KL divergence $\mathcal{L}_{kl}$. 
The loop-free nature of the architecture allows the decoder blocks run in parallel across various time steps, thus saving much inference time.
}
\vspace{-8mm}
\label{pipline_baselines}
\end{center}
\end{figure}

Based on these findings, we propose the \textit{first} Time-independent Unified Encoder (\ourmethod) design for the UNet in distilled student models, which inherits the same structure as the SD model. Specifically, the one-pass scheme means that the student UNet encoder is used only once, while the student UNet decoder processes across multiple time steps with shared weights (\cref{pipline_baselines}d) when distilling latent diffusion models.
\ourmethod achieves \textit{loop-free image generation}, eliminating the need for iterative noisy latent processing by sharing encoder features across multi-step decoders. This property supports decoder parallelization across diverse time steps, reducing inference time. Additionally, by using the decoder in multiple steps to extract more explicit semantic information from different time steps, \ourmethod produces more diverse and realistic results than prior methods in the one-step generation setup, with a similar time cost to previous one-step T2I models, such as LCM~\cite{luo2023latent}, SD-Turbo~\cite{sauer2023adversarial}, and SwiftBrushv2~\cite{dao2024swiftbrushv2}.

To further eliminate the need for large image datasets (either real or synthesized by the teacher model) as required by previous methods~\cite{nguyen2023swiftbrush}, we introduce a Kullback–Leibler (KL) divergence term to regularize the UNet's noise prediction, ensuring that it remains close to a standard normal distribution. While prior methods primarily focus on a diffusion distillation loss, they often overlook the importance of maintaining the UNet's Gaussian output prior. 
We empirically find that the KL divergence helps to enhance both perceptual realism and generation diversity. In summary, our work makes the following contributions:
\begin{itemize}[leftmargin=*]
    \item We introduce the \textit{first} Time-independent Unified Encoder (\ourmethod) architecture, which is a loop-free distillation approach and eliminates the need for iterative noisy latent processing while maintaining high sampling fidelity with a time cost comparable to previous one-step methods.
     \item We propose a novel diffusion distillation method that enables parallel sampling in diffusion models. Additionally, we incorporate KL divergence to regulate the output distribution of the student network, enhancing both generation diversity and perceptual realism.
     \item In both qualitative and quantitative experiments, our method outperforms strong baselines such as LCM\cite{luo2023latent}, SD-Turbo\cite{sauer2023adversarial}, and SwiftBrushv2~\cite{dao2024swiftbrushv2}, producing more diverse and realistic generative results.
\end{itemize}

\section{Related Work}
\label{sec:related_work}

Based on whether the acceleration methods require additional training, 
they can be categorized into 
training-based acceleration and training-free acceleration.

\subsection{Training-Free Acceleration}

The first main method of training-free acceleration optimizes the \textit{sample efficiency}~\cite{zheng2023_fastsampling,pokle_deep_equilibrium,geng2023onestep}. The sampling process can be performed by solving reverse SDEs or ODEs,  leading to extensive research on improved numerical solvers, including DDIM~\cite{song2021ddim}, DPM~\cite{lu2022dpm}, and DPM++~\cite{lu2022dpm++}.
Another key branch of training-free methods focuses on improving \textit{structural efficiency} during each sampling step, that includes network pruning~\cite{fang2024structural,su2024f3_pruning,zhang2024laptop_diff}, quantization~\cite{li2023q_diffusion,chen2024q_dit,shang2023post_quant}, parallelization~\cite{chen2024asyncdiff,li2024distrifusion,wang2024pipefusion} and token reduction~\cite{bolya2023token_merging,kim2024token_fusion,lou2024token_caching}.
More recently, a wide variety of methods utilize the \textit{cache mechanisms} to achieve training-free acceleration. These methods temporarily store information that can be reused to speed up computations. For example, DeepCache~\cite{ma2023deepcache} reduces redundant computations in Stable Diffusion by reusing intermediate features of low-resolution layers in the UNet. Faster Diffusion~\cite{li2024faster} accelerates the sampling process of diffusion models by caching UNet encoder features across timesteps, allowing the model to skip encoder computations at certain steps. Similar approaches have been applied to DiT-based T2I diffusion models, such as $\Delta$-DiT~\cite{chen2024delta_dit} and FORA~\cite{selvaraju2024fora}.

However, training-free methods~\cite{ma2024learning2cache,huang2024harmonica} generally apply the same caching solution to all tokens and still require multiple inference steps. This often leads to significant degradation in generation quality and poorer time complexity compared to training-based approaches.

\subsection{Training-Based Acceleration}
Several attempts have been made to accelerate the sampling process of diffusion models by introducing additional training beyond the base diffusion model.
One of the most representative methods is the \textit{Consistency Models}~\cite{song2023consistency,luo2023latent}.
Following works~\cite{zheng2024TCD,wang2024phased_CM} further improve the performance of these consistency models. 

More recently, \textit{distillation} techniques~\cite{hinton2014distilling,yin2024onestep_dmd,luo2024diff_instruct} have been applied to diffusion models, allowing faster training of student models by pre-trained teachers~\cite{sauer2023adversarial,nguyen2023swiftbrush,gu2023boot,liu2023instaflow}. Early studies~\cite{salimans2022progressive, meng2023distillation} used progressive distillation to gradually reduce the sampling steps required for student diffusion models. However, the slow sampling process of the pre-trained teacher limits the training efficiency.
To address this, recent works~\cite{gu2023boot,sauer2023adversarial,nguyen2023swiftbrush} have proposed various bootstrapping techniques. For example, Boot~\cite{gu2023boot} employs bootstrapping between two consecutive sampling steps, enabling image-free distillation. SD-Turbo~\cite{sauer2023adversarial} introduces a discriminator combined with a score distillation loss to improve performance. 
Most of these methods depend on image-text pair datasets for training, requiring substantial data alignment between visual and textual features. In contrast, SwiftBrush~\cite{nguyen2023swiftbrush} adapts variational score distillation.
SwiftBrush achieves the first \textit{image-free} training, 
avoiding the need for paired datasets. 
SwiftBrushv2~\cite{dao2024swiftbrushv2} further augments this method by merging two distilled student models.

However, existing training-based methods largely overlook feature similarities and redundant computations within UNet encoder-decoder (Enc-Dec) architectures, which have been widely explored in training-free acceleration techniques. Unlike conventional time-dependent Enc-Dec designs in popular T2I diffusion models and training-based accelerations, we propose the first Time-independent Unified Encoder (\ourmethod) model design for T2I diffusion models. In this approach, 
the encoder features are shared with the decoder across various time steps.

\begin{figure*}[tb]
\begin{center}
\includegraphics[width=0.88\linewidth]{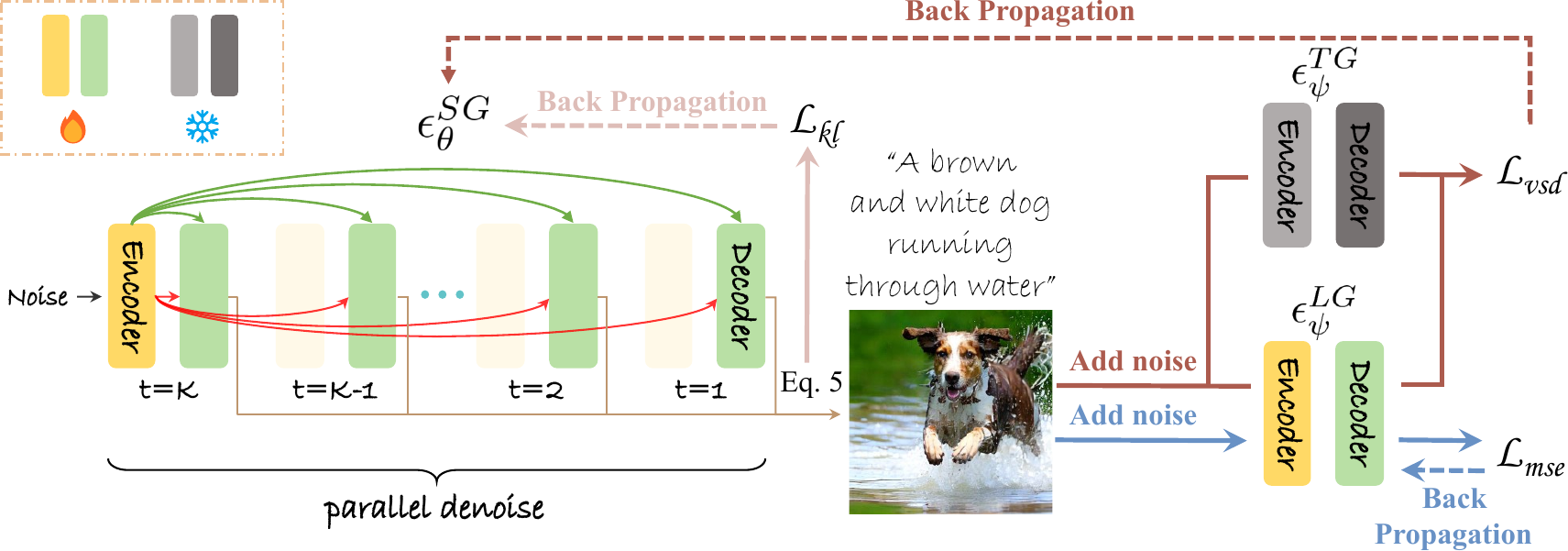}
\vspace{-4mm}
\caption{Loop-free distillation based on our Time-independent Unified Encoder (\ourmethod) architecture.  Our framework is composed of three networks: a student generator $\epsilon^{SG}_{\theta}$, a pretrained teacher SD generator $\epsilon^{TG}_{\psi}$ and a SD-LoRA generator $\epsilon^{LG}_{\psi}$. 
The encoder of $\epsilon^{SG}_{\theta}$ is only used once while passing through the decoder for \textit{K} time-steps. 
The student network is updated with both variational score distillation loss $\mathcal{L}_{vsd}$ and KL divergence loss $\mathcal{L}_{kl}$. 
We also optimize the SD-LoRA $\epsilon^{LG}_{\psi}$ by diffusion loss. The SD-LoRA teacher and the student model are updated alternately, following the previous methods~\cite{wang2023prolificdreamer,nguyen2023swiftbrush}.}
\vspace{-8mm}
\label{pipline}
\end{center}
\end{figure*}

\vspace{-2mm}
\section{Method}
\label{sec:method}
\vspace{-1mm}
Our goal is to distill a pretrained T2I diffusion model, typically a latent diffusion model (\cref{diffusion_vsd}), into a fast, loop-free student generator. The student model should maintain both the high quality and diversity of generated images.
Rather than adopting the conventional Time-dependent UNet encoder-decoder design, we introduce the Time-independent Unified Encoder (\ourmethod), which utilizes the UNet encoder in a single pass while the decoder operates in multiple time steps. Additionally, we incorporate a Kullback–Leibler (KL) divergence term to ensure that the UNet output adheres to a normal Gaussian distribution, preserving the fundamental diffusion model property that added noise follows a normal distribution. The procedure for our proposed method is introduced in \cref{loop_free} and \cref{pipline}.

\subsection{Preliminaries}
\label{diffusion_vsd}

\minisection{Latent Diffusion Model.}
The denoising network $\epsilon_{\psi}$ takes as input  a text  ${y}$, a latent code $\boldsymbol{z_t}$ and a time $t$ embedding to predict noise,
resulting in a latent $\boldsymbol{z_{t-1}}$. Using the DDIM scheduler~\cite{song2021ddim}, the formula is: 
\begin{equation}\label{eq:ddim_sampling}
\resizebox{1.\linewidth}{!}{$
\boldsymbol{z_{t-1}} = \sqrt{\frac{\alpha_{t-1}}{\alpha_t}}\boldsymbol{z_{t}} + \sqrt{\alpha_{t-1}}\left(\sqrt{\frac{1}{\alpha_{t-1}}-1}-\sqrt{\frac{1}{\alpha_t}-1}\right) \cdot \epsilon_{\psi}(\boldsymbol{z_{t}},t,{y})$},
\end{equation}
where $\alpha_t$ is a predefined scalar function at time-step $t$ ($t =T,..., 1$). The denoising network is commonly a UNet~\cite{ronneberger2015unet} consisting of an encoder ${E}$ and a decoder ${D}$.

\minisection{Variational Score Distillation (VSD).}
Score Distillation
Sampling (SDS) is an optimization method that involves distilling pretrained diffusion models $\epsilon_{\psi}(x_t, t, y)$ for text-to-3D generation~\cite{poole2023dreamfusion,wang2023score}. It optimizes the parameters of the generator $g(\theta)$, associated with a specific text $y$, using a loss function with an approximated gradient:
\vspace{-2mm}
\begin{equation}\scriptstyle
\nabla_\theta \mathcal{L}_{SDS} = \mathbb{E}_{t, \epsilon} \Bigl[w(t)(\epsilon_{\psi}(x_t, t, y) - \epsilon)\frac{\partial g(\theta)}{\partial \theta}\Bigr]
\end{equation}
where $w(t)$ is a time-dependent weighting function and 
$\epsilon {\sim} \mathcal{N}(0, I)$, $x_t = \alpha_t g(\theta) + \sigma_t \epsilon, t {\sim} \mathcal{U}(0.02T, 0.98T)$.

However, SDS suffers from over-saturation and over-smoothing when generating images with both diversity and fidelity. VSD~\cite{wang2023prolificdreamer} attributes this problem to the use of a single-point Dirac distribution as the variational distribution. To avoid this, VSD optimizes for the whole distribution through relevant gradients:
\vspace{-2mm}
\begin{equation}\label{eq:loss_vsd}
\begin{split}\scriptstyle
\nabla_\theta \mathcal{L}_{VSD} =  
 \mathbb{E}_{t, \epsilon} 
 \Bigl[
 &w(t)(\epsilon_{\psi}(x_t, t, y) 
 - \epsilon_{\phi}(x_t, t, y)\color{black})\frac{\partial g(\theta)}{\partial \theta} 
 \Bigr]
\end{split}
\end{equation}

VSD employs LoRA~\cite{hu2021lora} of the pretrained model $\epsilon_{\psi}(x_t, t, y)$ to parameterize $\epsilon_{\phi}(x_t, t, y)$. It is trained with the standard diffusion loss as the mean-square error:
\begin{equation}\label{eq:loss_lora}
    \min_{\epsilon_{\phi}} \mathbb{E}_{t, \epsilon} \underbrace{\Vert \epsilon_{\phi}(x_{t}, t, y) - \epsilon \Vert_2^2}_{\mathcal{L}_{mse}}
    \vspace{-1mm}
\end{equation}
To ensure that $\epsilon_{\phi}(x_t, t, y)$ aligns with the current distribution of the generator $g(\theta)$, VSD employs alternate optimization of $\epsilon_{\phi}$ and $\theta$.
Note that, for the sake of simplicity, we omit the angle-related variables required for text-to-3D.

VSD extends the variational formulation of SDS, providing an explanation and addressing issues that are observed with SDS.
Similar to SwiftBrush~\cite{nguyen2023swiftbrush}, we use VSD to distill a text-to-image pretrained model.

\subsection{\ourmethod: Time-independent Unified Encoder}
\label{loop_free}

Inspired by the observation in Faster Diffusion~\cite{li2024faster} that encoder features are sharing similarity across adjacent inference steps,
we explore the correlation between image generation quality and encoder feature similarity in SD model~\cite{rombach2022high} in~\cref{fig:feat_similarity}. 
With inference steps over 15 steps, high-quality images are associated with high encoder feature similarity. Below this threshold, the image generation quality deteriorates while the encoder feature exhibits variations.
This phenomenon also exists in distilled few-step diffusion models, as observed in~\cref{fig:features}.
Based on these observations, we propose the Time-independent Unified Encoder (\ourmethod) model design for distilling T2I diffusion models. The one-pass scheme means that the student UNet encoder is used only once, while the student UNet decoder processes across multiple time steps (\cref{pipline_baselines}d) when distilling latent diffusion models. \cref{pipline} shows our knowledge distillation framework.

Our distillation framework is composed of three networks: a pretrained teacher generator $\epsilon^{TG}_{\psi}$, an SD-LoRA generator $\epsilon^{LG}_{\psi}$ and an \ourmethod student generator $\epsilon^{SG}_{\theta}$.
Each generator consists of an encoder $E$ and a decoder $D$.  The student generator $\epsilon^{SG}_{\theta}$ takes as input a random noise $\boldsymbol{\epsilon}$, a text ${y}$, time-step $K$ and $t$, outputs a latent representation {$\epsilon^{SG}_{\theta}(\boldsymbol{\epsilon}, K, t, {y})$} ($t = K, ..., 1$).   $K$ is the input time step of the student generator $\epsilon^{SG}_{\theta}$  encoder, and   $t$ ($t = K,..., 1$)  is the input time-step of the student generator $\epsilon^{SG}_{\theta}$  decoder. 
The~\cref{eq:ddim_sampling} can be rewritten as: 
\begin{equation}\label{eq:our_sampling}
\resizebox{0.999\linewidth}{!}{
$\boldsymbol{z_{t-1}} = \sqrt{\frac{\alpha_{t-1}}{\alpha_t}}\boldsymbol{z_{t}} + \sqrt{\alpha_{t-1}}\left(\sqrt{\frac{1}{\alpha_{t-1}}-1}-\sqrt{\frac{1}{\alpha_t}-1}\right) \cdot \epsilon^{SG}_{\theta}(\boldsymbol{\epsilon},K, t, {y})$}.
\end{equation}
By this means, our method \ourmethod also adheres to the original ODE trajectory~\cite{song2021ddim,lu2022dpm}.
Furthermore, at time step $t$ ($t=K\text{-}1,...,1$), the decoder inputs do not depend on the encoder outputs at time step $t$. Instead, it relies on the encoder output at the key $K$ time step, as shown in \cref{pipline}. This allows us to perform \textit{parallel denoising} to generate the latent representation $\boldsymbol{z_0}$.
The final latent representation $\boldsymbol{z_0}$ is fed into both the pretrained teacher generator $\epsilon^{TG}_{\psi}$ and the SD-LoRA generator $\epsilon^{LG}_{\psi}$.
We update the student generator using \cref{eq:loss_vsd} in combination with \cref{eq:KL}, and then update the SD-LoRA generator with \cref{eq:loss_lora}. 
With all these techniques, our approach \ourmethod involves an alternating strategy for training both the student generator and SD-LoRA.

With the proposed strategy, we have two advantages. (1) When using the UNet encoder at a single step and the UNet decoder across multiple time-steps,
we are able to predict the latent noise in parallel,  achieving \textit{loop-free} image generation and significantly reducing inference time.
(2) Although we only use the encoder in a single step, we preserve multistep UNet decoders, which play a crucial role in generating high-quality images~\cite{zhang2023adding,tang2023emergent,tumanyan2023plug,zhang2023tale}. Therefore, we are able to produce more diverse and realistic images.

\minisection{KL divergence loss.}
Current distillation methods~\cite{luo2023latent,sauer2023adversarial}
commonly begin with initializing the student model $\epsilon^{SG}_{\theta}$  with the pretrained weight of the T2I teacher model.
This initialization indicates that the UNet output prior follows a Gaussian distribution (see~\cref{eq:loss_lora}). 
To fully follow the pretrained model to preserve this characteristic, one intuitive way is to take advantage of the diffusion loss (see~\cref{eq:loss_lora}) to train the student model. 
However, we expect to avoid access to a significant amount of images (real data or synthetically generated) for training. 
Instead, we introduce a KL divergence to regulate the UNet output distribution during the training of student generator $\epsilon^{SG}_{\theta}$: 
\begin{equation}
\mathcal{L}_{\text{KL}} =  [ \mathcal{D}_{\text{KL}}(\epsilon^{SG}_{\theta}(\boldsymbol{\epsilon}, T_1, t, \boldsymbol{y}) || \;\mathcal{N}(0,I)) ],
\vspace{-1mm}
\label{eq:KL}
\end{equation}
where $\mathcal{D}_{\text{KL}}(p||q)=-\int  p(z) \log\frac{p(z)}{q(z)}dz $.   

Finally, we train the student model $\epsilon^{SG}_{\theta}$ using the VSD (\cref{eq:loss_vsd}) and KL divergence (\cref{eq:KL}). Meanwhile, we use the denoising loss (\cref{eq:loss_lora}) to update the SD-LoRA model $\epsilon^{LG}_{\psi}$.
Note that we alternately train both the student model $\epsilon^{SG}_{\theta}$ and the SD-LoRA network $\epsilon^{LG}_{\psi}$ while freezing the text-to-image teacher $\epsilon^{TG}_{\psi}$.

\begin{table*}[tb]
\renewcommand{\arraystretch}{0.07}
\tabcolsep=0.04cm
\centering
\resizebox{0.999\linewidth}{!}{
\begin{tabular}{c|c|c|c|c|c|c|c|c|c|c|c|c|c|c|c|c|c|c}
\toprule
Dataset & {\multirow{4}{*}{\makecell{Base\\Model}}}& {\multirow{4}{*}{Step}} & {\multirow{4}{*}{Param}} & \multicolumn{5}{c|}{COCO2014-30K}  & \multicolumn{5}{c|}{\centering COCO2017-5K} & \multicolumn{2}{c|}{\makecell{Inference$\downarrow$}} & \multicolumn{2}{c|}{\makecell{Training Data}} & \multirow{4}{*}{\makecell{A100\\Days$\downarrow$}}\\
\cmidrule{1-1}\cmidrule{5-14}\cmidrule{15-18}
\diagbox{Method}{Metrics} & & & & {FID$\downarrow$} & {CLIP$\uparrow$} & {Precision$\uparrow$} & {Recall$\uparrow$} & {F1$\uparrow$}& {FID$\downarrow$} & {CLIP$\uparrow$} & {Precision$\uparrow$} & {Recall$\uparrow$} & {F1$\uparrow$}& \makecell{Time\\(ms)} & \makecell{Memory\\(GB)} & Size$\downarrow$ & \makecell{Image\\Free}\\
\midrule\midrule
SD1.5~\cite{rombach2022high} (\textit{cfg}=7.5)$^\dag$ & -- & 50 & 860M & 16.08 & 0.325 & 0.717 & 0.527 & 0.607 & 23.39 & 0.326 & 0.776 & 0.587& 0.668 & 2503.0 & 4.04 & 5B& \xmarkg & 4783\\
SD1.5~\cite{rombach2022high} (\textit{cfg}=4.5)$^\dag$ & -- & 50 & 860M & 9.90 & 0.322 & 0.727 & 0.585 & 0.648 & 19.87 & 0.323 & 0.764 & 0.649 & 0.702 & 2503.0 & 4.04 & 5B& \xmarkg & 4783\\
SD2.1~\cite{rombach2022high} (\textit{cfg}=7.5)$^\dag$ & -- & 50 & 865M & 16.10 & 0.328 & 0.723 & 0.489 & 0.583 & 25.40 & 0.328 & 0.769 & 0.561 & 0.649 & 2244.2 & 3.89 & 5B & \xmarkg & 8332\\
SD2.1~\cite{rombach2022high} (\textit{cfg}=4.5)$^\dag$ & -- & 50 & 865M & 12.22 & 0.325 & 0.734 & 0.526 & 0.614 & 22.24 & 0.298 & 0.788 & 0.606 & 0.685 & 2244.2 & 3.89 & 5B& \xmarkg & 8332\\
\midrule
GigaGAN~\cite{kang2023scaling}$^*$ & GAN & 1 & 1.0B & 9.24 & 0.325 & 0.724 & 0.547 & 0.623 & -- & -- & -- & -- & -- & -- & -- & 2.7B & \xmarkg & 6250\\
\midrule\midrule
InstaFlow~\cite{liu2023instaflow}$^\dag$ & \multirow{2}{*}{SD1.5} & 1 & 0.9B & \underline{13.78} & 0.288 & 0.654 & \underline{0.521} &  \underline{0.580} & \textbf{19.00} &0.293 & 0.729 & \underline{0.613} &  \underline{0.666} & 111.3 & 3.99 & 3.2M & \xmarkg & 183.2\\
LCM~\cite{luo2023latent}$^\dag$ & & 1 & 860M & 132.09 & 0.230 & 0.109 & 0.194 & 0.140 & 143.73 & 0.229 & 0.118 & 0.291 & 0.168 &236.2 & 5.88 & 12M&\xmarkg & 1.3\\
\cmidrule{2-2}
{SD-Turbo}~\cite{sauer2023adversarial}$^\dag$ & \multirow{3.5}{*}{SD2.1} & 1 & 865M & 19.51 & \underline{0.331} & \underline{0.758} & 0.458 & 0.571 & 29.35 & \underline{0.331} & \underline{0.786} & 0.445 & 0.568 &140.0 & 3.86 &unk.&\xmarkg & unk.\\
SwiftBrush~\cite{nguyen2023swiftbrush}$^\dag$ & & 1 & 865M & 17.20 & 0.301 & 0.672 & 0.458 & 0.545 & 27.18 & 0.314 & 0.729 & 0.527 & 0.612 & 95.0 & 3.85 & 1.4M & \cmarkg & 4.1\\
SwiftBrushv2~\cite{dao2024swiftbrushv2}$^\ddag$ & & 1 & 865M & 15.98 & {0.326} & \textbf{0.782} & 0.457 & 0.577 & 26.28 & 0.326 & \textbf{0.816} & 0.543 & 0.652 & 139.6 & 4.91& 1.4M& \cmarkg & 24.1\\
\midrule
LCM~\cite{luo2023latent}$^\dag$ & SD1.5 & 4 & 860M & 23.21 & 0.262 & 0.666 & 0.346 & 0.455 & 40.37 & 0.303 & 0.713 & 0.460 & 0.559 &592.3 & 5.88 & 12M &\xmarkg & 1.3\\
{SD-Turbo}~\cite{sauer2023adversarial}$^\dag$ & SD2.1 & 4 & 865M & 16.14 & \textbf{0.335} & 0.633 & 0.394 & 0.468& 26.14 & \textbf{0.335} & 0.694 & 0.375 & 0.487 & 272.2 & 3.86 &unk.&\xmarkg & unk.\\
\midrule
{Ours} & SD2.1 & 1 & 865M & \textbf{13.09} & {0.313} & 0.634 & \textbf{0.622} & \textbf{0.628} & \underline{23.11}  & 0.313 & 0.697 & \textbf{0.668} & \textbf{0.682} & 164.7 & 4.98 & 1.4M & \cmarkg & 3.9 \\
\bottomrule
\end{tabular}
}
\vspace{-3mm}
\caption{Comparison of our distillation method against other works. 
 Inference Time (ms) and Memory (GB). 
$^\dag$ indicates that we report results using the provided official code and pretrained models. 
$^\ddag$ denotes that we re-implemented the work and are providing the scores.
$^*$ indicates that we report results using the provided generated images.
``unk.'' denotes unknown.
The best and second-best scores are highlighted in \textbf{bold} and \underline{underlined}, respectively, with both the parameter count and training data size being below the billion level.
}
\vspace{-2mm}
\label{tab:fid_clip}
\end{table*}

\begin{figure*}[tb]
\begin{center}
\includegraphics[width=0.979\linewidth]{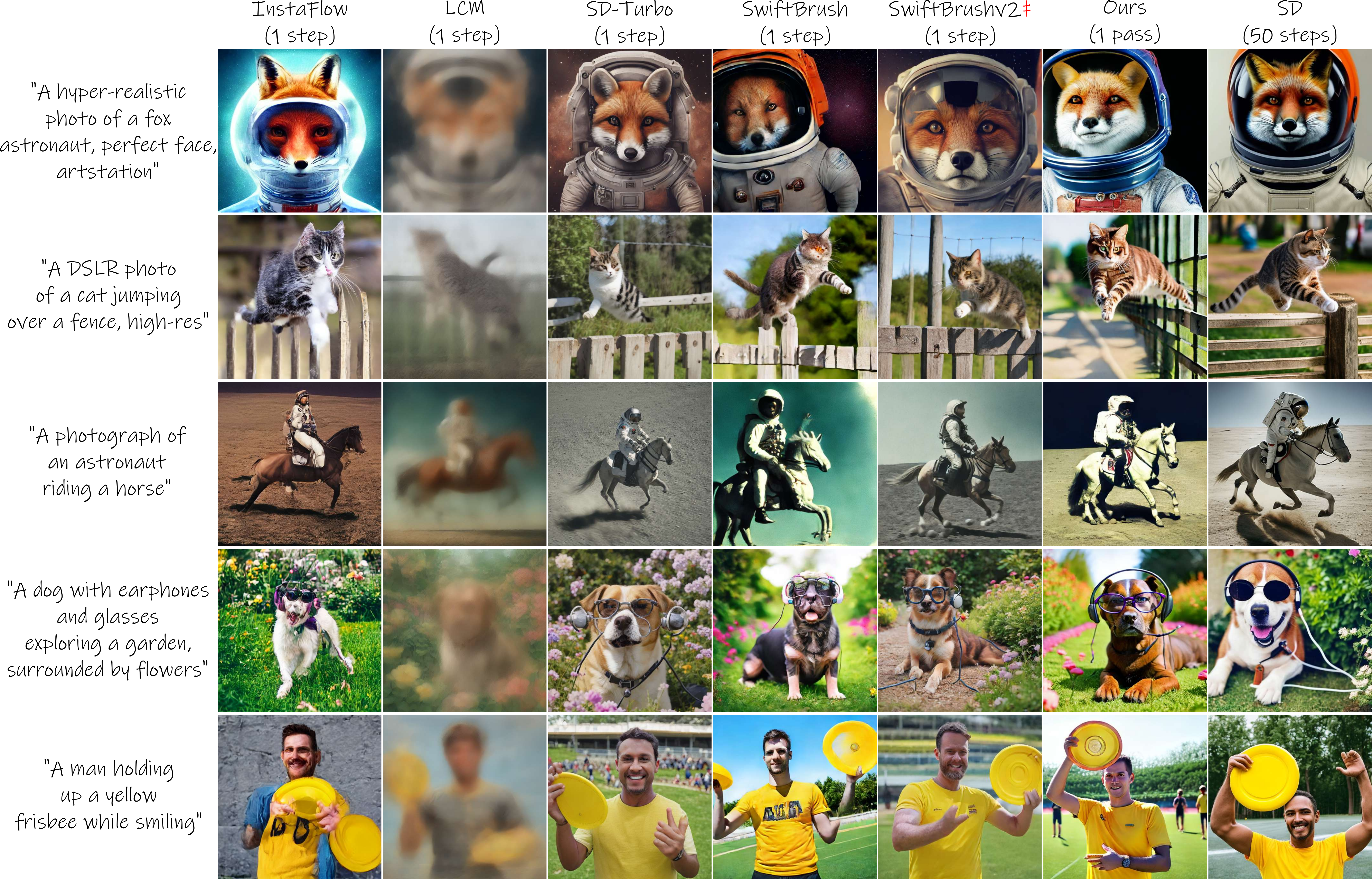}
\vspace{-3mm}
\caption{{Qualitative comparison to state-of-the-art few-step distillation methods.} $^\ddag$ denotes that we re-implemented the work.
Our model outperforms all other few-step samplers in terms of quality within one-pass, approaching the performance of the SD model with 50 steps.
}
\label{fig:qualitative_comparison}
\end{center}
\vspace{-8mm}
\end{figure*}

\vspace{-2mm}
\section{Experiments}
\vspace{-1mm}
\minisection{Evaluation Datasets and Metrics.} We compare our method \ourmethod with the following baselines: 
Instaflow~\cite{liu2023instaflow}, LCM~\cite{luo2023latent},  
SwiftBrush~\cite{nguyen2023swiftbrush}, SwiftBrushv2~\cite{dao2024swiftbrushv2} and SD-Turbo~\cite{sauer2023adversarial} across the zero-shot text-to-image benchmark COCO 2014 and 2017~\cite{lin2014microsoft}.
In COCO 2014, we randomly select 30K prompts as the conventional evaluation protocol~\cite{sauer2023stylegant, kang2023scaling, saharia2022photorealistic, rombach2022high}, and feed them into the diffusion model to obtain 30K generated images. 
In COCO 2017, we obtain 5K generated images using the provided 5K prompts. We use the Fréchet Inception Distance (FID)~\cite{heusel2017gans} metric to assess the visual quality of the generated images and the CLIPScore(CLIP)~\cite{hessel2021clipscore} to measure the consistency between the image content and the text prompt. Here, we use ViT-B/32 as the backbone to evaluate CLIPScore.
We also use Precision and Recall~\cite{kynkaanniemi2019improved} to quantify fidelity and diversity of generated images.
Additionally, we compute Density and Coverage metrics~\cite{naeem2020reliable}, along with FID, to assess the diversity of the generated images on the AFHQ~\cite{choi2020stargan}, CelebA-HQ~\cite{karras2017progressive} datasets and the prompt datasets DrawBench~\cite{saharia2022photorealistic} and PartiPrompts~\cite{yu2022scaling}.

\minisection{Implementation Details.} Similar to prior works~\cite{luo2023latent,sauer2023adversarial,nguyen2023swiftbrush}, we use the same network SD2.1-base as the teacher model, and initialize all student parameters using a pretrained teacher model. During the training process for the SD-LoRA generator, we apply a learning rate of 1e-3, a LoRA rank of 64, and an alpha value of 108, following SwiftBrush~\cite{nguyen2023swiftbrush}.
Simultaneously, the learning rate for the student generator is configured at 1e-6, incorporating the exponential moving average (EMA). We set $K=4$, and employ the Adam optimizer~\cite{kingma2017adam} to train both the student generator and SD-LoRA generator. %
In this context, we leverage 1.4M captions sourced from the extensive text-image dataset JourneyDB~\cite{pan2023journeydb}.
See more detailed implementation information in \blue{Appendix A}.

\vspace{-1mm}
\subsection{Quantitative and qualitative results}
\label{results}

\begin{figure}[t]
    \centering
    \includegraphics[width=0.999\linewidth, trim=0 0 0 10, clip]{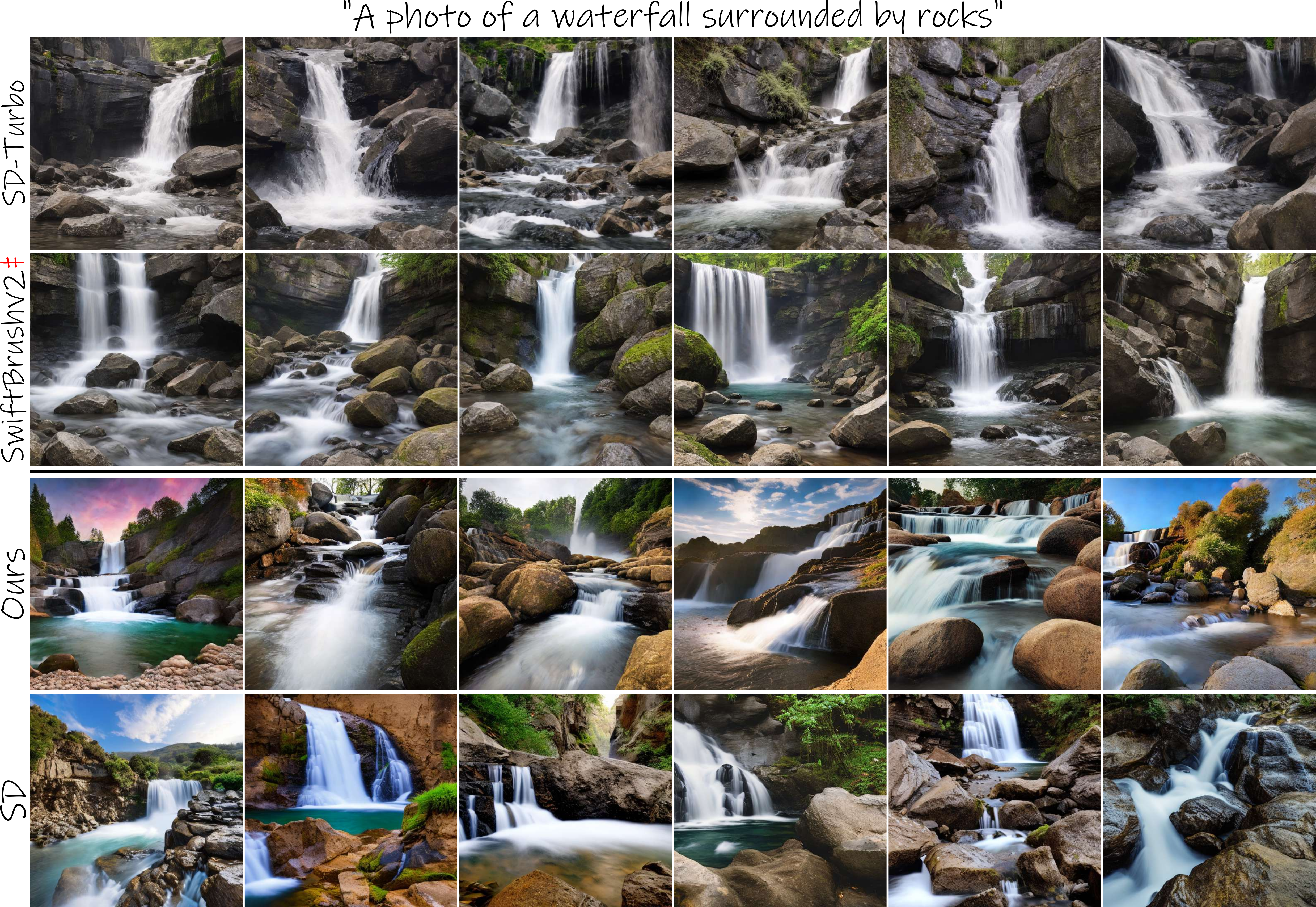}
    \vspace{-6mm}
    \caption{Both SD-Turbo and SwiftBrushv2 tend to generate results with similar scenery and style when given the same prompt, resulting in a lack of diversity. $^\ddag$ denotes that we re-implemented the work.}
    \vspace{-6mm}
      \label{fig:diversity}
\end{figure}

We conduct the comparison with state-of-the-art one-step distillation methods, including both image-dependent~\cite{liu2023instaflow,luo2023latent,sauer2023adversarial} and image-free ~\cite{nguyen2023swiftbrush,dao2024swiftbrushv2} techniques. As shown in~\cref{tab:fid_clip} (the last row),
despite being trained solely on text data, our approach \ourmethod achieves the best and second-best FID scores in two benchmarks and demonstrated similar performance in terms of CLIPScore score compared to SwiftBrushv1/v2.  
For both the inference time and the memory usage, LCM is nearly twice as much as ours, while SD-Turbo and SwiftBrushv1/v2  have a slight advantage over ours.
Since our approach use  encoder once, and does the decoder across \textit{K} time-steps (i.e., \textit{K}=4 in~\cref{tab:fid_clip} (the last row)), to ensure a more equitable comparison, we further evaluate our results against both LCM and SD-Turbo in 4-steps sampling (\cref{tab:fid_clip} (the eleventh and twelfth rows)).
Our approach outperforms LCM and SwiftBrush across all evaluation metrics, except for Precision, including FID, CLIPScore, and Recall. Furthermore, our method obtains comparable performance to SD-Turbo with 4-steps while  using less  inference time.

\cref{fig:qualitative_comparison} shows the generation of distillation methods with various steps.
Our result outperforms those of all other one-step generators in terms of quality. 
Our approach approaches the performance of the SD model with 50 steps.
We observe that both SD-Turbo and SwiftBrushv2 tend to generate images with limited  diversity (\cref{fig:diversity} (the first and second rows)), while they generate high-quality results.  Obviously, we are able to produce more realism and diverse results (\cref{fig:diversity} (the third row), and close to the ones of SD (\cref{fig:diversity} (the last row)), indicating our advantage over the baselines.
See \blue{Appendix B} for additional results.

\begin{figure}[t]
\begin{center}
\centerline{\includegraphics[width=\linewidth]{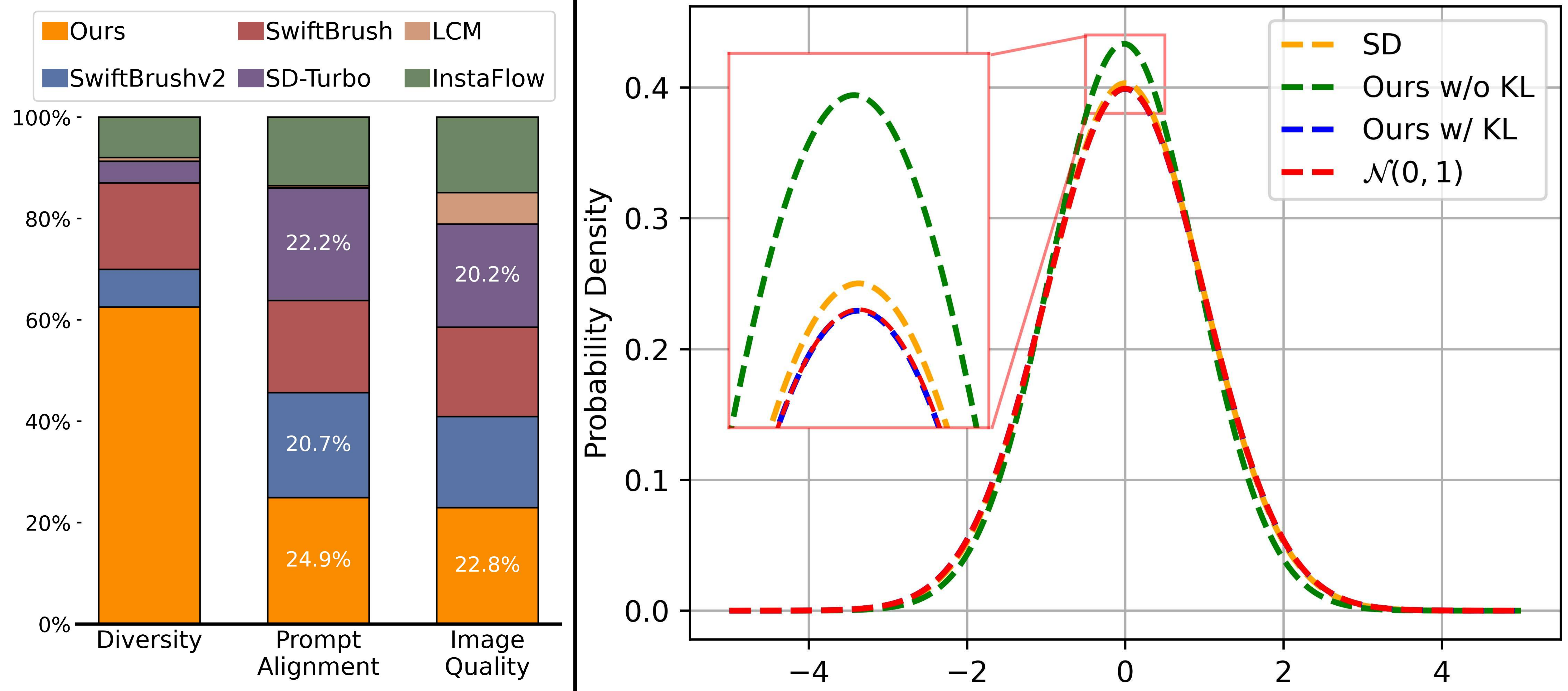}}
\vspace{-3mm}
\caption{User Study (left). Comparison of the predicted noise distribution between SD and our method (right).}
\vspace{-11mm}
\label{userstudy}
\end{center}
\end{figure}

\begin{table*}[tb]
\renewcommand{\arraystretch}{0.08}
\tabcolsep=0.01cm
\centering
\resizebox{1.0\linewidth}{!}{
\begin{tabular}{c|c|c|c|c|c|c|c|c|c|c|c|c|c|c|c}
\toprule
Dataset & \multirow{4}{*}{\makecell{Base\\Model}} & \multirow{4}{*}{Step} & \multicolumn{3}{c|}{AFHQ} & \multicolumn{3}{c|}{CelebA-HQ} & \multicolumn{3}{c|}{DrawBench}& \multicolumn{3}{c|}{PartiPrompts}  & {\makecell{Training\\Data}} \\
\cmidrule{1-1} \cmidrule{4-15} \cmidrule{16-16}
\diagbox{Method}{Metrics} & &	& FID$\downarrow$ & Density$\uparrow$ & Coverage$\uparrow$ & FID$\downarrow$ & Density$\uparrow$ & Coverage$\uparrow$& FID$\downarrow$ & Density$\uparrow$ & Coverage$\uparrow$& FID$\downarrow$ & Density$\uparrow$ & Coverage$\uparrow$ &  \makecell{Image\\Free}\\
\midrule\midrule
SD1.5~\cite{rombach2022high} (\textit{cfg}=7.5)$^\dag$ & -- & 50 & 47.16 & 0.066 & 0.030 & 93.94 & 0.053 & 0.013 & 11.95 & 0.510 & 0.622 & 7.36 & 0.730 & 0.887  & \xmarkg \\
SD2.1~\cite{rombach2022high} (\textit{cfg}=7.5)$^\dag$ & -- & 50 & 51.67 & 0.053 & 0.022 & 89.57 & 0.018 & 0.013 & 0 & 1 & 1 & 0 & 1 & 1 & \xmarkg \\
\midrule\midrule
InstaFlow~\cite{liu2023instaflow}$^\dag$ & \multirow{2}{*}{SD1.5} & 1 & \textbf{51.97} & 0.058 & \underline{0.029} & 131.99 & 0.026 & 0.007 & 25.08 & 0.223 & 0.337 & 17.64 & 0.457 & 0.670  & \xmarkg \\
LCM~\cite{luo2023latent}$^\dag$ &  & 1  & 155.63 & 0.012 & 0.033 & 165.74 & 0.001 & 0.004 & 120.98 & 0.058 & 0.014 & 95.65 & 0.095 & 0.072  & \xmarkg\\
\cmidrule{2-2}
SD-Turbo~\cite{sauer2023adversarial}$^\dag$ & \multirow{3.5}{*}{SD2.1} & 1 & 77.75 & \textbf{0.142} & 0.033 & 146.22 & 0.047 & 0.006 & 25.75 & 0.597 & 0.488 & 17.40 & 0.770 & 0.775  & \xmarkg\\
SwiftBrush~\cite{nguyen2023swiftbrush}$^\dag$ &  & 1 & 67.60 & 0.039 & 0.014 &  144.03  & 0.014 & 0.002 & 21.48 & 0.402 & 0.441 & \underline{14.43} & 0.579 & 0.737  & \cmarkg \\
SwiftBrushv2~\cite{dao2024swiftbrushv2}$^\ddag$ &  & 1 & 64.99 & \underline{0.110} & 0.025 & 131.89 & \underline{0.055} & 0.012 & \textbf{18.57} & \underline{0.682} & \underline{0.597} & \textbf{11.32} & \underline{0.850} & \textbf{0.865}  & \cmarkg \\
\midrule
LCM~\cite{luo2023latent}$^\dag$ & SD1.5 & 4  & 78.00 & 0.054 & {0.008} & \underline{122.44} & 0.045 & \underline{0.045} & 46.23 & 0.183 & 0.187 & 26.84 & 0.512 & 0.575 & \xmarkg \\
SD-Turbo~\cite{sauer2023adversarial}$^\dag$ & SD2.1 & 4 & 77.23 & 0.011 & 0.005 & 193.08 & 0.013 & 0.001 & 27.80 & 0.281 & 0.371 & 22.84 & 0.500 & 0.648  & \xmarkg \\
\midrule
Ours & SD2.1 & 1 & \underline{54.48} & {0.068} & \textbf{0.071} & \textbf{116.82} & \textbf{0.116} & \textbf{0.068} & \underline{21.10} &\textbf{0.685} & \textbf{0.616} & {16.28} & \textbf{0.852} & \underline{0.840} & \cmarkg \\
\bottomrule
\end{tabular}
}
\vspace{-3mm}
\caption{\small Quantitative comparison of our distillation method with other approaches based on FID, Density, and Coverage metrics to assess diversity. $^\dag$ indicates that we report results using the provided official code and pretrained models. 
$^\ddag$ denotes that we re-implemented the work and are providing the scores. The best and second-best numbers are marked with \textbf{bold} and \underline{underlined} respectively.}
\vspace{-6mm}
\label{tab:dns_cvg}
\end{table*}

We also report the quantitative results to evaluate the \textit{density} and \textit{coverage} metrics~\cite{naeem2020reliable}.  
We use 200 prompts from DrawBench, for each prompt we generate 100 images by using difference seeds. Similarly, we use 1632 prompts from PartiPrompts, generating 10 images per prompt. During the evaluation, we take the images synthesized by SD2.1 as the ground truth.
For the real image we use both AFHQ~\cite{choi2020stargan} and  CelebA-HQ~\cite{karras2017progressive}. 
As reported in \cref{tab:dns_cvg}, compared to all baselines,  our method achieves the best scores in terms of Density and Coverage on all three datasets, except for the Density in \textit{AFHQ} and Coverage in \textit{PartiPrompts}. 
These results indicate that we produce images with a high degree of diversity.

\minisection{User Study.}
We conducted a user study, as shown in \cref{userstudy} (left), and asked participants to select their preferred results (image quality, prompt alignment, and diversity). We compared 54 users (30 image sets/user) using multiple choices. 
Experimental results indicate that our method  has a significant results,  indicating our large advantage over the baselines. Furthermore, our method achieves better results comparing to the baseline methods in terms of \textit{image quality} and \textit{prompt alignment}.

\subsection{Additional Analysis}

\minisection{The impact of KL diverse regularization.} We regularize the student UNet output to close a Gaussian noise, aiming to use fully the pretrained model property.  To evaluate the effectiveness of KL diverse regularization, we analyze the distribution of the UNet decoder output values  and fit a curve to it.  As depicted in \cref{userstudy} (right),  without $\mathcal{L}_{\text{KL}}$  our approach exhibits noticeable deviations from the Gaussian distribution (green and red curves), whereas incorporating $\mathcal{L}_{\text{KL}}$ the UNet output curve is more closer to a Gaussian distribution (blue and red curves), resulting in the generation of high-quality images.
\cref{tab:kl_loss_abl} further reports the quantitative comparison of FID and CLIPScore, demonstrating that our method, combined with KL diverse regularization, exhibits a significant improvement.

\begin{table}[t]
\renewcommand{\arraystretch}{0.2}
\tabcolsep=0.01cm
        \centering
\resizebox{0.99\linewidth}{!}{
    \begin{tabular}{c|c|c|c|c|c|c}
        \toprule
        \diagbox{Method}{Metrics}  		  & FID$\downarrow$          & CLIP$\uparrow$ & {Precision$\uparrow$} & {Recall$\uparrow$} & Density$\uparrow$ & Coverage$\uparrow$ \\
        \midrule
        Ours w/o $\mathcal{L}_{\text{KL}}$	 & 14.90 & 0.311 & 0.608 & 0.621 & 0.617 & 0.631\\
        Ours & 13.09 & 0.313 & 0.634 & 0.622 & 0.683 & 0.652 \\
        \bottomrule
    \end{tabular}
     }
    \vspace{-3mm}
\caption{Ablation study by quantitative evaluation for the KL diverse regularization $\mathcal{L}_{KL}$.}
\label{tab:kl_loss_abl}
    \vspace{-5mm}
\end{table}

\begin{figure}[t]
    \centering
    \includegraphics[width=0.999\linewidth, trim=0 0 0 10, clip]{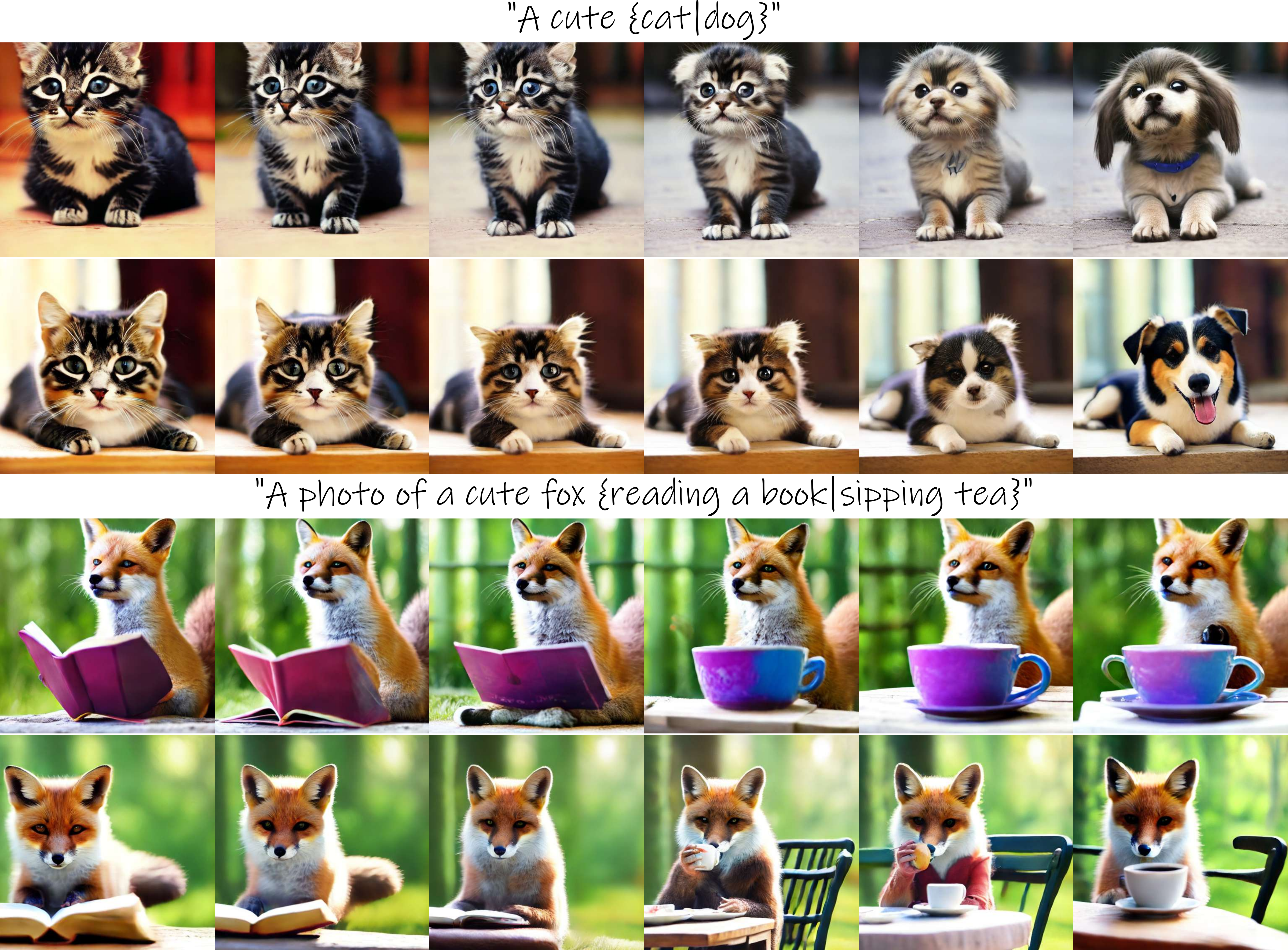}
    \vspace{-6mm}
    \caption{Interpolation between text prompts. For example, interpolation between two animals in the first two rows with the prompts ``... cat'' (left) and ``... dog'' (right). Interpolation between two actions in last two rows with the prompts ``... reading a book'' (left) and ``... sipping tea'' (right).}
    \vspace{-6mm}
      \label{fig:inter}
\end{figure}

\minisection{Text prompt Interpolation.}
~\cref{fig:inter} showcases the interpolation process, where each row maintains a fixed input noise while the text embedding transitions smoothly between two distinct text prompts. 
Even when presented with previously unseen interpolated text embeddings, our model demonstrates remarkable semantic consistency, producing images that align continuously with the interpolated prompts and maintain high-quality generative outputs.

\minisection{Hyperparameter \textit{K}.}
The parameter \textit{K} decides how many decoders to be used, which influences the quality and diversity. For both LCM and SD-Turbo, they normally use 4-steps to generate satisfactory results, which inspires us to set \textit{K=4} as the main experimental setup. Although we do not traverse all possible \textit{K} values, we still achieve better results (e.g., FID) with  \textit{K}= 4.  

\begin{figure}[t]
\begin{center}
\centerline{\includegraphics[width=0.999\linewidth]{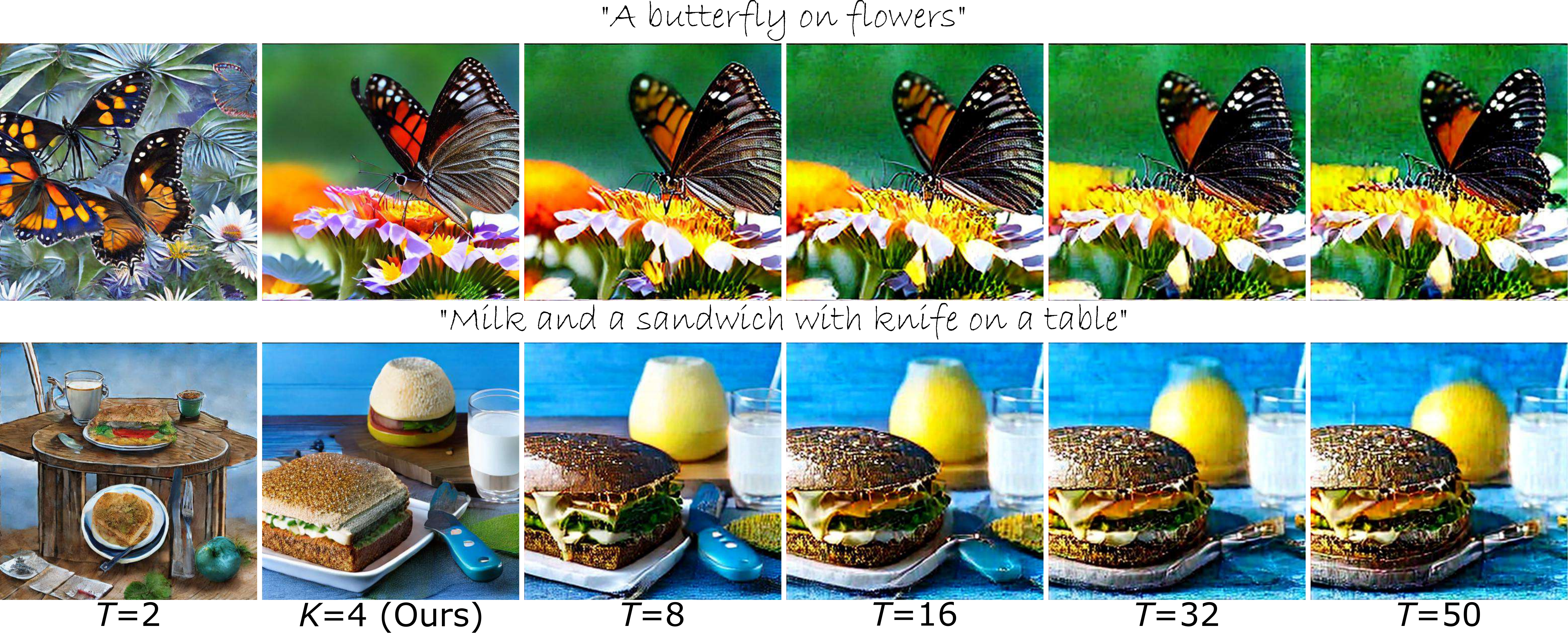}}
\vspace{-3mm}
\caption{Results of different time-steps with \textit{K}=4 trained student model. The well-trained student model with time-steps \textit{K}=4 (the second column) serves as a good starting point for students set at other time-steps (e.g., 2, 8, 16, 32, and 50).
}
\vspace{-12mm}
\label{diff_K}
\end{center}
\end{figure}

\minisection{Diverse sampling steps during inference.} Although the student network is optimized with both VSD and KL-divergence losses without the denoising loss, we are still able to sample with various time steps. 
As examples in~\cref{diff_K} sampled with the DDIM scheduler~\cite{song2021ddim} using different time-steps (e.g., 2, 8, 16, 32, and 50), the generated images still contain significantly semantic information, while it loses the image quality. 
This experiment demonstrates that our method greatly inherits the capabilities of the pretrained model.
We expect that fine-tuning the well-trained student model with \(K=4\) for several iterations will produce superior results at larger time-steps.

\vspace{-2mm}
\section{Conclusions}\vspace{-1mm}
In this work, we introduce the Time-independent Unified Encoder (\ourmethod), a major advancement in T2I diffusion distillation that improves inference speed and generation quality. Our findings reveal that decoders effectively capture richer semantic information, while encoders can be shared across multiple decoders from different time steps. This inspires a novel one-pass scheme that enables efficient encoder utilization, achieving loop-free image generation and reduced time complexity. Additionally, incorporating Kullback–Leibler (KL) divergence enhances the noise prediction process, ensuring high perceptual realism and diversity in outputs. Extensive experiments show that \ourmethod surpasses state-of-the-art methods like LCM, SD-Turbo, and SwiftBrushv1/v2, yielding more diverse and realistic results.

\section*{Acknowledgements}
This work was supported by NSFC (NO. 62225604), Youth Foundation (62202243), and Shenzhen Science and Technology Program (JCYJ20240813114237048).
We acknowledge the support of the project PID2022-143257NB-I00, funded by the Spanish Government through MCIN/AEI/10.13039/501100011033 and FEDER, and the Generalitat de Catalunya CERCA Program.
We acknowledge ``Science and Technology Yongjiang 2035'' key technology breakthrough plan project (2024Z120).
Computation is supported by the Supercomputing Center of Nankai University (NKSC).

{
    \small
    \bibliographystyle{ieeenat_fullname}
    \bibliography{longstrings,main}

\begin{thebibliography}{86}
\providecommand{\natexlab}[1]{#1}
\providecommand{\url}[1]{\texttt{#1}}
\expandafter\ifx\csname urlstyle\endcsname\relax
  \providecommand{\doi}[1]{doi: #1}\else
  \providecommand{\doi}{doi: \begingroup \urlstyle{rm}\Url}\fi

\bibitem[Bolya and Hoffman(2023)]{bolya2023token_merging}
Daniel Bolya and Judy Hoffman.
\newblock Token merging for fast stable diffusion.
\newblock In \emph{Proceedings of the IEEE/CVF Conference on Computer Vision and Pattern Recognition}, pages 4598--4602, 2023.

\bibitem[Butt et~al.(2024)Butt, Wang, Vazquez-Corral, and van~de Weijer]{butt2025colorpeel}
Muhammad~Atif Butt, Kai Wang, Javier Vazquez-Corral, and Joost van~de Weijer.
\newblock Colorpeel: Color prompt learning with diffusion models via color and shape disentanglement.
\newblock In \emph{ECCV}, 2024.

\bibitem[Chen et~al.(2025)Chen, Meng, Tang, Ma, Jiang, Wang, Wang, and Zhu]{chen2024q_dit}
Lei Chen, Yuan Meng, Chen Tang, Xinzhu Ma, Jingyan Jiang, Xin Wang, Zhi Wang, and Wenwu Zhu.
\newblock Q-dit: Accurate post-training quantization for diffusion transformers.
\newblock \emph{CVPR}, 2025.

\bibitem[Chen et~al.(2024{\natexlab{a}})Chen, Shen, Ye, Cao, Tu, Bouganis, Zhao, and Chen]{chen2024delta_dit}
Pengtao Chen, Mingzhu Shen, Peng Ye, Jianjian Cao, Chongjun Tu, Christos-Savvas Bouganis, Yiren Zhao, and Tao Chen.
\newblock Delta-dit: A training-free acceleration method tailored for diffusion transformers.
\newblock \emph{arXiv preprint arXiv:2406.01125}, 2024{\natexlab{a}}.

\bibitem[Chen et~al.(2024{\natexlab{b}})Chen, Ma, Fang, Tan, and Wang]{chen2024asyncdiff}
Zigeng Chen, Xinyin Ma, Gongfan Fang, Zhenxiong Tan, and Xinchao Wang.
\newblock Asyncdiff: Parallelizing diffusion models by asynchronous denoising.
\newblock \emph{NeurIPS}, 2024{\natexlab{b}}.

\bibitem[Choi et~al.(2020)Choi, Uh, Yoo, and Ha]{choi2020stargan}
Yunjey Choi, Youngjung Uh, Jaejun Yoo, and Jung-Woo Ha.
\newblock Stargan v2: Diverse image synthesis for multiple domains.
\newblock In \emph{Proceedings of the IEEE/CVF conference on computer vision and pattern recognition}, pages 8188--8197, 2020.

\bibitem[Dao et~al.(2024)Dao, Nguyen, Le, Vu, Nguyen, Pham, and Tran]{dao2024swiftbrushv2}
Trung Dao, Thuan~Hoang Nguyen, Thanh Le, Duc Vu, Khoi Nguyen, Cuong Pham, and Anh Tran.
\newblock Swiftbrush v2: Make your one-step diffusion model better than its teacher.
\newblock \emph{ECCV}, 2024.

\bibitem[DeepFloyd(2023)]{IF}
DeepFloyd.
\newblock {DeepFloyd IF}.
\newblock \url{https://www.deepfloyd.ai/deepfloyd-if}, 2023.

\bibitem[Dhariwal and Nichol(2021)]{dhariwal2021diffusionbeatgans}
Prafulla Dhariwal and Alexander Nichol.
\newblock Diffusion models beat gans on image synthesis.
\newblock \emph{Advances in neural information processing systems}, 34:\penalty0 8780--8794, 2021.

\bibitem[Fang et~al.(2024)Fang, Ma, and Wang]{fang2024structural}
Gongfan Fang, Xinyin Ma, and Xinchao Wang.
\newblock Structural pruning for diffusion models.
\newblock \emph{Advances in neural information processing systems}, 36, 2024.

\bibitem[Geng et~al.(2023)Geng, Pokle, and Kolter]{geng2023onestep}
Zhengyang Geng, Ashwini Pokle, and J~Zico Kolter.
\newblock One-step diffusion distillation via deep equilibrium models.
\newblock In \emph{Thirty-seventh Conference on Neural Information Processing Systems}, 2023.

\bibitem[Geyer et~al.(2024)Geyer, Bar-Tal, Bagon, and Dekel]{tokenflow2023}
Michal Geyer, Omer Bar-Tal, Shai Bagon, and Tali Dekel.
\newblock Tokenflow: Consistent diffusion features for consistent video editing.
\newblock \emph{ICLR}, 2024.

\bibitem[Gu et~al.(2023{\natexlab{a}})Gu, Wang, Zhao, Fu, Xiong, Liu, Zhang, Zhang, Zhang, Jung, and Wang]{gu2023photoswap}
Jing Gu, Yilin Wang, Nanxuan Zhao, Tsu-Jui Fu, Wei Xiong, Qing Liu, Zhifei Zhang, He Zhang, Jianming Zhang, HyunJoon Jung, and Xin~Eric Wang.
\newblock Photoswap: Personalized subject swapping in images, 2023{\natexlab{a}}.

\bibitem[Gu et~al.(2023{\natexlab{b}})Gu, Zhai, Zhang, Liu, and Susskind]{gu2023boot}
Jiatao Gu, Shuangfei Zhai, Yizhe Zhang, Lingjie Liu, and Josh Susskind.
\newblock Boot: Data-free distillation of denoising diffusion models with bootstrapping.
\newblock \emph{International Conference on Machine Learning}, 2023{\natexlab{b}}.

\bibitem[Hertz et~al.(2023)Hertz, Mokady, Tenenbaum, Aberman, Pritch, and Cohen-Or]{hertz2022prompt}
Amir Hertz, Ron Mokady, Jay Tenenbaum, Kfir Aberman, Yael Pritch, and Daniel Cohen-Or.
\newblock Prompt-to-prompt image editing with cross attention control.
\newblock \emph{ICLR}, 2023.

\bibitem[Hessel et~al.(2021)Hessel, Holtzman, Forbes, Le~Bras, and Choi]{hessel2021clipscore}
Jack Hessel, Ari Holtzman, Maxwell Forbes, Ronan Le~Bras, and Yejin Choi.
\newblock Clipscore: A reference-free evaluation metric for image captioning.
\newblock In \emph{Proceedings of the 2021 Conference on Empirical Methods in Natural Language Processing}, pages 7514--7528, 2021.

\bibitem[Heusel et~al.(2017)Heusel, Ramsauer, Unterthiner, Nessler, and Hochreiter]{heusel2017gans}
Martin Heusel, Hubert Ramsauer, Thomas Unterthiner, Bernhard Nessler, and Sepp Hochreiter.
\newblock Gans trained by a two time-scale update rule converge to a local nash equilibrium.
\newblock In \emph{NeurIPS}, pages 6626--6637, 2017.

\bibitem[Hinton et~al.(2014)Hinton, Vinyals, and Dean]{hinton2014distilling}
Geoffrey Hinton, Oriol Vinyals, and Jeff Dean.
\newblock {Distilling the Knowledge in a Neural Network}.
\newblock \emph{NIPS Deep Learning Workshop}, 2014.

\bibitem[Ho et~al.(2020)Ho, Jain, and Abbeel]{ho2020ddpm}
Jonathan Ho, Ajay Jain, and Pieter Abbeel.
\newblock Denoising diffusion probabilistic models.
\newblock \emph{Advances in Neural Information Processing Systems}, 33:\penalty0 6840--6851, 2020.

\bibitem[Hu et~al.(2022)Hu, Shen, Wallis, Allen-Zhu, Li, Wang, Wang, and Chen]{hu2021lora}
Edward~J Hu, Yelong Shen, Phillip Wallis, Zeyuan Allen-Zhu, Yuanzhi Li, Shean Wang, Lu Wang, and Weizhu Chen.
\newblock Lora: Low-rank adaptation of large language models.
\newblock \emph{ICLR}, 2022.

\bibitem[Hu et~al.(2024)Hu, Li, van~de Weijer, Gao, Khan, Yang, Cheng, Wang, and Wang]{hu2024tome}
Taihang Hu, Linxuan Li, Joost van~de Weijer, Hongcheng Gao, Fahad~Shahbaz Khan, Jian Yang, Mingming Cheng, Kai Wang, and Yaxing Wang.
\newblock Token merging for training-free semantic binding in text-to-image synthesis.
\newblock In \emph{NeurIPS}, 2024.

\bibitem[Huang et~al.(2024)Huang, Wang, Gong, Liu, Zhang, Guo, Liu, and Zhang]{huang2024harmonica}
Yushi Huang, Zining Wang, Ruihao Gong, Jing Liu, Xinjie Zhang, Jinyang Guo, Xianglong Liu, and Jun Zhang.
\newblock Harmonica: Harmonizing training and inference for better feature cache in diffusion transformer acceleration.
\newblock \emph{arXiv preprint arXiv:2410.01723}, 2024.

\bibitem[Kang and Park(2020)]{Kang2020ContraGANCL}
Mingu Kang and Jaesik Park.
\newblock Contragan: Contrastive learning for conditional image generation.
\newblock \emph{NeurIPS}, 2020.

\bibitem[Kang et~al.(2021)Kang, Shim, Cho, and Park]{Kang2021RebootingAA}
Minguk Kang, Woohyeon Shim, Minsu Cho, and Jaesik Park.
\newblock Rebooting acgan: Auxiliary classifier gans with stable training.
\newblock In \emph{Neural Information Processing Systems}, 2021.

\bibitem[Kang et~al.(2023{\natexlab{a}})Kang, Shin, and Park]{kang2023studiogan}
Minguk Kang, Joonghyuk Shin, and Jaesik Park.
\newblock Studiogan: a taxonomy and benchmark of gans for image synthesis.
\newblock \emph{IEEE Transactions on Pattern Analysis and Machine Intelligence}, 2023{\natexlab{a}}.

\bibitem[Kang et~al.(2023{\natexlab{b}})Kang, Zhu, Zhang, Park, Shechtman, Paris, and Park]{kang2023scaling}
Minguk Kang, Jun-Yan Zhu, Richard Zhang, Jaesik Park, Eli Shechtman, Sylvain Paris, and Taesung Park.
\newblock {Scaling up GANs for Text-to-Image Synthesis}.
\newblock \emph{CVPR}, 2023{\natexlab{b}}.

\bibitem[Karras et~al.(2018)Karras, Aila, Laine, and Lehtinen]{karras2017progressive}
Tero Karras, Timo Aila, Samuli Laine, and Jaakko Lehtinen.
\newblock Progressive growing of gans for improved quality, stability, and variation.
\newblock \emph{ICLR}, 2018.

\bibitem[Khachatryan et~al.(2023)Khachatryan, Movsisyan, Tadevosyan, Henschel, Wang, Navasardyan, and Shi]{text2video-zero}
Levon Khachatryan, Andranik Movsisyan, Vahram Tadevosyan, Roberto Henschel, Zhangyang Wang, Shant Navasardyan, and Humphrey Shi.
\newblock Text2video-zero: Text-to-image diffusion models are zero-shot video generators.
\newblock \emph{ICCV}, 2023.

\bibitem[Kim et~al.(2024)Kim, Gao, Hsu, Shen, and Jin]{kim2024token_fusion}
Minchul Kim, Shangqian Gao, Yen-Chang Hsu, Yilin Shen, and Hongxia Jin.
\newblock Token fusion: Bridging the gap between token pruning and token merging.
\newblock In \emph{Proceedings of the IEEE/CVF Winter Conference on Applications of Computer Vision}, pages 1383--1392, 2024.

\bibitem[Kingma and Ba(2015)]{kingma2017adam}
Diederik~P. Kingma and Jimmy Ba.
\newblock {Adam: A Method for Stochastic Optimization}.
\newblock \emph{ICLR}, 2015.

\bibitem[Kumari et~al.(2023)Kumari, Zhang, Zhang, Shechtman, and Zhu]{kumari2023multi}
Nupur Kumari, Bingliang Zhang, Richard Zhang, Eli Shechtman, and Jun-Yan Zhu.
\newblock Multi-concept customization of text-to-image diffusion.
\newblock In \emph{Proceedings of the IEEE/CVF Conference on Computer Vision and Pattern Recognition}, pages 1931--1941, 2023.

\bibitem[Kynk{\"a}{\"a}nniemi et~al.(2019)Kynk{\"a}{\"a}nniemi, Karras, Laine, Lehtinen, and Aila]{kynkaanniemi2019improved}
Tuomas Kynk{\"a}{\"a}nniemi, Tero Karras, Samuli Laine, Jaakko Lehtinen, and Timo Aila.
\newblock Improved precision and recall metric for assessing generative models.
\newblock \emph{Advances in neural information processing systems}, 32, 2019.

\bibitem[Li et~al.(2024{\natexlab{a}})Li, Cai, Cao, Zhang, Cai, Bai, Jia, Li, and Han]{li2024distrifusion}
Muyang Li, Tianle Cai, Jiaxin Cao, Qinsheng Zhang, Han Cai, Junjie Bai, Yangqing Jia, Kai Li, and Song Han.
\newblock Distrifusion: Distributed parallel inference for high-resolution diffusion models.
\newblock In \emph{Proceedings of the IEEE/CVF Conference on Computer Vision and Pattern Recognition}, pages 7183--7193, 2024{\natexlab{a}}.

\bibitem[Li et~al.(2024{\natexlab{b}})Li, Hu, van~de Weijer, Khan, Liu, Li, Yang, Wang, Cheng, and Yang]{li2024faster}
Senmao Li, Taihang Hu, Joost van~de Weijer, Fahad~Shahbaz Khan, Tao Liu, Linxuan Li, Shiqi Yang, Yaxing Wang, Ming-Ming Cheng, and Jian Yang.
\newblock Faster diffusion: Rethinking the role of unet encoder in diffusion models.
\newblock In \emph{Advances in Neural Information Processing Systems}, 2024{\natexlab{b}}.

\bibitem[Li et~al.(2024{\natexlab{c}})Li, van~de Weijer, Hu, Khan, Hou, Wang, and Yang]{li2023stylediffusion}
Senmao Li, Joost van~de Weijer, Taihang Hu, Fahad~Shahbaz Khan, Qibin Hou, Yaxing Wang, and Jian Yang.
\newblock Stylediffusion: Prompt-embedding inversion for text-based editing.
\newblock \emph{Computational Visual Media Conference}, 2024{\natexlab{c}}.

\bibitem[Li et~al.(2023)Li, Liu, Lian, Yang, Dong, Kang, Zhang, and Keutzer]{li2023q_diffusion}
Xiuyu Li, Yijiang Liu, Long Lian, Huanrui Yang, Zhen Dong, Daniel Kang, Shanghang Zhang, and Kurt Keutzer.
\newblock Q-diffusion: Quantizing diffusion models.
\newblock In \emph{Proceedings of the IEEE/CVF International Conference on Computer Vision}, pages 17535--17545, 2023.

\bibitem[Lin et~al.(2024)Lin, Wang, and Yang]{lin2024sdxl_lightning}
Shanchuan Lin, Anran Wang, and Xiao Yang.
\newblock Sdxl-lightning: Progressive adversarial diffusion distillation.
\newblock \emph{arXiv preprint arXiv:2402.13929}, 2024.

\bibitem[Lin et~al.(2014)Lin, Maire, Belongie, Hays, Perona, Ramanan, Doll{\'a}r, and Zitnick]{lin2014microsoft}
Tsung-Yi Lin, Michael Maire, Serge Belongie, James Hays, Pietro Perona, Deva Ramanan, Piotr Doll{\'a}r, and C~Lawrence Zitnick.
\newblock Microsoft coco: Common objects in context.
\newblock In \emph{Computer Vision--ECCV 2014: 13th European Conference, Zurich, Switzerland, September 6-12, 2014, Proceedings, Part V 13}, pages 740--755. Springer, 2014.

\bibitem[Liu et~al.(2025)Liu, Wang, Li, van~de Weijer, Khan, Yang, Wang, Yang, and Cheng]{liu2025onepromptonestory}
Tao Liu, Kai Wang, Senmao Li, Joost van~de Weijer, Fahad~Shahbaz Khan, Shiqi Yang, Yaxing Wang, Jian Yang, and Ming-Ming Cheng.
\newblock One-prompt-one-story: Free-lunch consistent text-to-image generation using a single prompt.
\newblock \emph{CVPR}, 2025.

\bibitem[Liu et~al.(2024)Liu, Zhang, Ma, Peng, and Liu]{liu2023instaflow}
Xingchao Liu, Xiwen Zhang, Jianzhu Ma, Jian Peng, and Qiang Liu.
\newblock Instaflow: One step is enough for high-quality diffusion-based text-to-image generation.
\newblock \emph{ICLR}, 2024.

\bibitem[Lou et~al.(2024)Lou, Luo, Liu, Li, Ding, Hu, Cao, Li, and Ma]{lou2024token_caching}
Jinming Lou, Wenyang Luo, Yufan Liu, Bing Li, Xinmiao Ding, Weiming Hu, Jiajiong Cao, Yuming Li, and Chenguang Ma.
\newblock Token caching for diffusion transformer acceleration.
\newblock \emph{arXiv preprint arXiv:2409.18523}, 2024.

\bibitem[Lu et~al.(2022{\natexlab{a}})Lu, Zhou, Bao, Chen, Li, and Zhu]{lu2022dpm}
Cheng Lu, Yuhao Zhou, Fan Bao, Jianfei Chen, Chongxuan Li, and Jun Zhu.
\newblock Dpm-solver: A fast ode solver for diffusion probabilistic model sampling in around 10 steps.
\newblock \emph{Advances in Neural Information Processing Systems}, 35:\penalty0 5775--5787, 2022{\natexlab{a}}.

\bibitem[Lu et~al.(2022{\natexlab{b}})Lu, Zhou, Bao, Chen, Li, and Zhu]{lu2022dpm++}
Cheng Lu, Yuhao Zhou, Fan Bao, Jianfei Chen, Chongxuan Li, and Jun Zhu.
\newblock Dpm-solver++: Fast solver for guided sampling of diffusion probabilistic models.
\newblock \emph{arXiv preprint arXiv:2211.01095}, 2022{\natexlab{b}}.

\bibitem[Luo et~al.(2023{\natexlab{a}})Luo, Tan, Huang, Li, and Zhao]{luo2023latent}
Simian Luo, Yiqin Tan, Longbo Huang, Jian Li, and Hang Zhao.
\newblock Latent consistency models: Synthesizing high-resolution images with few-step inference.
\newblock \emph{arXiv preprint arXiv:2310.04378}, 2023{\natexlab{a}}.

\bibitem[Luo et~al.(2023{\natexlab{b}})Luo, Tan, Patil, Gu, von Platen, Passos, Huang, Li, and Zhao]{luo2023lcm_lora}
Simian Luo, Yiqin Tan, Suraj Patil, Daniel Gu, Patrick von Platen, Apolin{\'a}rio Passos, Longbo Huang, Jian Li, and Hang Zhao.
\newblock Lcm-lora: A universal stable-diffusion acceleration module.
\newblock \emph{arXiv preprint arXiv:2311.05556}, 2023{\natexlab{b}}.

\bibitem[Luo et~al.(2023{\natexlab{c}})Luo, Hu, Zhang, Sun, Li, and Zhang]{luo2024diff_instruct}
Weijian Luo, Tianyang Hu, Shifeng Zhang, Jiacheng Sun, Zhenguo Li, and Zhihua Zhang.
\newblock Diff-instruct: A universal approach for transferring knowledge from pre-trained diffusion models.
\newblock \emph{NeurIPS}, 36, 2023{\natexlab{c}}.

\bibitem[Luo et~al.(2023{\natexlab{d}})Luo, Chen, Zhang, Huang, Wang, Shen, Zhao, Zhou, and Tan]{luo2023videofusion}
Zhengxiong Luo, Dayou Chen, Yingya Zhang, Yan Huang, Liang Wang, Yujun Shen, Deli Zhao, Jingren Zhou, and Tieniu Tan.
\newblock Videofusion: Decomposed diffusion models for high-quality video generation.
\newblock In \emph{Proceedings of the IEEE/CVF Conference on Computer Vision and Pattern Recognition}, pages 10209--10218, 2023{\natexlab{d}}.

\bibitem[Ma et~al.(2024{\natexlab{a}})Ma, Fang, Mi, and Wang]{ma2024learning2cache}
Xinyin Ma, Gongfan Fang, Michael~Bi Mi, and Xinchao Wang.
\newblock Learning-to-cache: Accelerating diffusion transformer via layer caching.
\newblock \emph{NeurIPS}, 2024{\natexlab{a}}.

\bibitem[Ma et~al.(2024{\natexlab{b}})Ma, Fang, and Wang]{ma2023deepcache}
Xinyin Ma, Gongfan Fang, and Xinchao Wang.
\newblock Deepcache: Accelerating diffusion models for free.
\newblock \emph{CVPR}, 2024{\natexlab{b}}.

\bibitem[Meng et~al.(2023)Meng, Rombach, Gao, Kingma, Ermon, Ho, and Salimans]{meng2023distillation}
Chenlin Meng, Robin Rombach, Ruiqi Gao, Diederik Kingma, Stefano Ermon, Jonathan Ho, and Tim Salimans.
\newblock On distillation of guided diffusion models.
\newblock In \emph{Proceedings of the IEEE/CVF Conference on Computer Vision and Pattern Recognition}, pages 14297--14306, 2023.

\bibitem[Mou et~al.(2023)Mou, Wang, Xie, Zhang, Qi, Shan, and Qie]{mou2023t2i}
Chong Mou, Xintao Wang, Liangbin Xie, Jian Zhang, Zhongang Qi, Ying Shan, and Xiaohu Qie.
\newblock T2i-adapter: Learning adapters to dig out more controllable ability for text-to-image diffusion models.
\newblock \emph{AAAI}, 2023.

\bibitem[Naeem et~al.(2020)Naeem, Oh, Uh, Choi, and Yoo]{naeem2020reliable}
Muhammad~Ferjad Naeem, Seong~Joon Oh, Youngjung Uh, Yunjey Choi, and Jaejun Yoo.
\newblock Reliable fidelity and diversity metrics for generative models.
\newblock In \emph{International Conference on Machine Learning}, pages 7176--7185. PMLR, 2020.

\bibitem[Nguyen and Tran(2024)]{nguyen2023swiftbrush}
Thuan~Hoang Nguyen and Anh Tran.
\newblock Swiftbrush: One-step text-to-image diffusion model with variational score distillation.
\newblock \emph{CVPR}, 2024.

\bibitem[Pan et~al.(2023)Pan, Sun, Ge, Li, Duan, Wu, Zhang, Zhou, Qin, Wang, Dai, Qiao, and Li]{pan2023journeydb}
Junting Pan, Keqiang Sun, Yuying Ge, Hao Li, Haodong Duan, Xiaoshi Wu, Renrui Zhang, Aojun Zhou, Zipeng Qin, Yi Wang, Jifeng Dai, Yu Qiao, and Hongsheng Li.
\newblock {JourneyDB: A Benchmark for Generative Image Understanding}.
\newblock \emph{NeurIPS}, 2023.

\bibitem[Pokle et~al.(2024)Pokle, Geng, and Kolter]{pokle_deep_equilibrium}
Ashwini Pokle, Zhengyang Geng, and Zico Kolter.
\newblock Deep equilibrium approaches to diffusion models.
\newblock In \emph{Proceedings of the 36th International Conference on Neural Information Processing Systems}, Red Hook, NY, USA, 2024. Curran Associates Inc.

\bibitem[Poole et~al.(2023)Poole, Jain, Barron, and Mildenhall]{poole2023dreamfusion}
Ben Poole, Ajay Jain, Jonathan~T. Barron, and Ben Mildenhall.
\newblock {DreamFusion: Text-to-3D using 2D Diffusion}.
\newblock \emph{ICLR}, 2023.

\bibitem[Ren et~al.(2024)Ren, Xia, Lu, Zhang, Wu, Xie, Wang, and Xiao]{ren2024hyper}
Yuxi Ren, Xin Xia, Yanzuo Lu, Jiacheng Zhang, Jie Wu, Pan Xie, Xing Wang, and Xuefeng Xiao.
\newblock Hyper-sd: Trajectory segmented consistency model for efficient image synthesis.
\newblock \emph{NeurIPS}, 2024.

\bibitem[Rombach et~al.(2022)Rombach, Blattmann, Lorenz, Esser, and Ommer]{rombach2022high}
Robin Rombach, Andreas Blattmann, Dominik Lorenz, Patrick Esser, and Bj{\"o}rn Ommer.
\newblock High-resolution image synthesis with latent diffusion models.
\newblock In \emph{Proceedings of the IEEE/CVF conference on computer vision and pattern recognition}, pages 10684--10695, 2022.

\bibitem[Ronneberger et~al.(2015)Ronneberger, Fischer, and Brox]{ronneberger2015unet}
Olaf Ronneberger, Philipp Fischer, and Thomas Brox.
\newblock U-net: Convolutional networks for biomedical image segmentation.
\newblock In \emph{Medical Image Computing and Computer-Assisted Intervention--MICCAI 2015: 18th International Conference, Munich, Germany, October 5-9, 2015, Proceedings, Part III 18}, pages 234--241. Springer, 2015.

\bibitem[Ruiz et~al.(2023)Ruiz, Li, Jampani, Pritch, Rubinstein, and Aberman]{ruiz2023dreambooth}
Nataniel Ruiz, Yuanzhen Li, Varun Jampani, Yael Pritch, Michael Rubinstein, and Kfir Aberman.
\newblock Dreambooth: Fine tuning text-to-image diffusion models for subject-driven generation.
\newblock In \emph{Proceedings of the IEEE/CVF Conference on Computer Vision and Pattern Recognition}, pages 22500--22510, 2023.

\bibitem[Saharia et~al.(2022)Saharia, Chan, Saxena, Li, Whang, Denton, Ghasemipour, Ayan, Mahdavi, Lopes, Salimans, Salimans, Ho, Fleet, and Norouzi]{saharia2022photorealistic}
Chitwan Saharia, William Chan, Saurabh Saxena, Lala Li, Jay Whang, Emily Denton, Seyed Kamyar~Seyed Ghasemipour, Burcu~Karagol Ayan, S~Sara Mahdavi, Rapha~Gontijo Lopes, Tim Salimans, Tim Salimans, Jonathan Ho, David~J Fleet, and Mohammad Norouzi.
\newblock Photorealistic text-to-image diffusion models with deep language understanding.
\newblock \emph{NeurIPS}, 2022.

\bibitem[Salimans and Ho(2022)]{salimans2022progressive}
Tim Salimans and Jonathan Ho.
\newblock Progressive distillation for fast sampling of diffusion models.
\newblock \emph{ICLR}, 2022.

\bibitem[Sauer et~al.(2023)Sauer, Karras, Laine, Geiger, and Aila]{sauer2023stylegant}
Axel Sauer, Tero Karras, Samuli Laine, Andreas Geiger, and Timo Aila.
\newblock {StyleGAN-T: Unlocking the Power of GANs for Fast Large-Scale Text-to-Image Synthesis}.
\newblock \emph{International Conference on Machine Learning}, 2023.

\bibitem[Sauer et~al.(2024)Sauer, Lorenz, Blattmann, and Rombach]{sauer2023adversarial}
Axel Sauer, Dominik Lorenz, Andreas Blattmann, and Robin Rombach.
\newblock Adversarial diffusion distillation.
\newblock \emph{ECCV}, 2024.

\bibitem[Selvaraju et~al.(2024)Selvaraju, Ding, Chen, Zharkov, and Liang]{selvaraju2024fora}
Pratheba Selvaraju, Tianyu Ding, Tianyi Chen, Ilya Zharkov, and Luming Liang.
\newblock Fora: Fast-forward caching in diffusion transformer acceleration.
\newblock \emph{arXiv preprint arXiv:2407.01425}, 2024.

\bibitem[Shang et~al.(2023)Shang, Yuan, Xie, Wu, and Yan]{shang2023post_quant}
Yuzhang Shang, Zhihang Yuan, Bin Xie, Bingzhe Wu, and Yan Yan.
\newblock Post-training quantization on diffusion models.
\newblock In \emph{Proceedings of the IEEE/CVF conference on computer vision and pattern recognition}, pages 1972--1981, 2023.

\bibitem[Song et~al.(2021{\natexlab{a}})Song, Meng, and Ermon]{song2021ddim}
Jiaming Song, Chenlin Meng, and Stefano Ermon.
\newblock Denoising diffusion implicit models.
\newblock In \emph{International Conference on Learning Representations}, 2021{\natexlab{a}}.

\bibitem[Song et~al.(2021{\natexlab{b}})Song, Sohl-Dickstein, Kingma, Kumar, Ermon, and Poole]{song2020score}
Yang Song, Jascha Sohl-Dickstein, Diederik~P Kingma, Abhishek Kumar, Stefano Ermon, and Ben Poole.
\newblock Score-based generative modeling through stochastic differential equations.
\newblock \emph{ICLR}, 2021{\natexlab{b}}.

\bibitem[Song et~al.(2023)Song, Dhariwal, Chen, and Sutskever]{song2023consistency}
Yang Song, Prafulla Dhariwal, Mark Chen, and Ilya Sutskever.
\newblock Consistency models.
\newblock In \emph{International Conference on Machine Learning}, pages 32211--32252. PMLR, 2023.

\bibitem[Su et~al.(2024)Su, Liu, Gao, and Song]{su2024f3_pruning}
Sitong Su, Jianzhi Liu, Lianli Gao, and Jingkuan Song.
\newblock F3-pruning: A training-free and generalized pruning strategy towards faster and finer text-to-video synthesis.
\newblock In \emph{Proceedings of the AAAI Conference on Artificial Intelligence}, pages 4961--4969, 2024.

\bibitem[Tang et~al.(2023)Tang, Jia, Wang, Phoo, and Hariharan]{tang2023emergent}
Luming Tang, Menglin Jia, Qianqian Wang, Cheng~Perng Phoo, and Bharath Hariharan.
\newblock Emergent correspondence from image diffusion.
\newblock \emph{NeurIPS}, 2023.

\bibitem[Tumanyan et~al.(2023)Tumanyan, Geyer, Bagon, and Dekel]{tumanyan2023plug}
Narek Tumanyan, Michal Geyer, Shai Bagon, and Tali Dekel.
\newblock Plug-and-play diffusion features for text-driven image-to-image translation.
\newblock In \emph{Proceedings of the IEEE/CVF Conference on Computer Vision and Pattern Recognition}, pages 1921--1930, 2023.

\bibitem[Wang et~al.(2024{\natexlab{a}})Wang, Huang, Bergman, Shen, Gao, Lingelbach, Sun, Bian, Song, Liu, et~al.]{wang2024phased_CM}
Fu-Yun Wang, Zhaoyang Huang, Alexander~William Bergman, Dazhong Shen, Peng Gao, Michael Lingelbach, Keqiang Sun, Weikang Bian, Guanglu Song, Yu Liu, et~al.
\newblock Phased consistency model.
\newblock \emph{NeurIPS}, 2024{\natexlab{a}}.

\bibitem[Wang et~al.(2023{\natexlab{a}})Wang, Du, Li, Yeh, and Shakhnarovich]{wang2023score}
Haochen Wang, Xiaodan Du, Jiahao Li, Raymond~A. Yeh, and Greg Shakhnarovich.
\newblock {Score Jacobian Chaining: Lifting Pretrained 2D Diffusion Models for 3D Generation}.
\newblock \emph{CVPR}, 2023{\natexlab{a}}.

\bibitem[Wang et~al.(2024{\natexlab{b}})Wang, Fang, Li, and Yang]{wang2024pipefusion}
Jiannan Wang, Jiarui Fang, Aoyu Li, and PengCheng Yang.
\newblock Pipefusion: Displaced patch pipeline parallelism for inference of diffusion transformer models.
\newblock \emph{arXiv preprint arXiv:2405.14430}, 2024{\natexlab{b}}.

\bibitem[Wang et~al.(2023{\natexlab{b}})Wang, Yang, Yang, Butt, and van~de Weijer]{kai2023DPL}
Kai Wang, Fei Yang, Shiqi Yang, Muhammad~Atif Butt, and Joost van~de Weijer.
\newblock Dynamic prompt learning: Addressing cross-attention leakage for text-based image editing.
\newblock \emph{NeurIPS}, 2023{\natexlab{b}}.

\bibitem[Wang et~al.(2025)Wang, Yang, Raducanu, and van~de Weijer]{wang2024mcti}
Kai Wang, Fei Yang, Bogdan Raducanu, and Joost van~de Weijer.
\newblock Multi-class textual-inversion secretly yields a semantic-agnostic classifier.
\newblock In \emph{Proceedings of the {IEEE} Workshop on Applications of Computer Vision}, 2025.

\bibitem[Wang et~al.(2023{\natexlab{c}})Wang, Lu, Wang, Bao, Li, Su, and Zhu]{wang2023prolificdreamer}
Zhengyi Wang, Cheng Lu, Yikai Wang, Fan Bao, Chongxuan Li, Hang Su, and Jun Zhu.
\newblock {ProlificDreamer: High-Fidelity and Diverse Text-to-3D Generation with Variational Score Distillation}.
\newblock \emph{NeurIPS}, 2023{\natexlab{c}}.

\bibitem[Wu et~al.(2023)Wu, Ge, Wang, Lei, Gu, Shi, Hsu, Shan, Qie, and Shou]{wu2023tune}
Jay~Zhangjie Wu, Yixiao Ge, Xintao Wang, Stan~Weixian Lei, Yuchao Gu, Yufei Shi, Wynne Hsu, Ying Shan, Xiaohu Qie, and Mike~Zheng Shou.
\newblock Tune-a-video: One-shot tuning of image diffusion models for text-to-video generation.
\newblock In \emph{Proceedings of the IEEE/CVF International Conference on Computer Vision}, pages 7623--7633, 2023.

\bibitem[Yin et~al.(2024)Yin, Gharbi, Zhang, Shechtman, Durand, Freeman, and Park]{yin2024onestep_dmd}
Tianwei Yin, Micha{\"e}l Gharbi, Richard Zhang, Eli Shechtman, Fredo Durand, William~T Freeman, and Taesung Park.
\newblock One-step diffusion with distribution matching distillation.
\newblock In \emph{CVPR}, pages 6613--6623, 2024.

\bibitem[Yu et~al.(2022)Yu, Xu, Koh, Luong, Baid, Wang, Vasudevan, Ku, Yang, Ayan, et~al.]{yu2022scaling}
Jiahui Yu, Yuanzhong Xu, Jing~Yu Koh, Thang Luong, Gunjan Baid, Zirui Wang, Vijay Vasudevan, Alexander Ku, Yinfei Yang, Burcu~Karagol Ayan, et~al.
\newblock Scaling autoregressive models for content-rich text-to-image generation.
\newblock \emph{arXiv preprint arXiv:2206.10789}, 2022.

\bibitem[Zhang et~al.(2024)Zhang, Li, Chen, Xie, and Lu]{zhang2024laptop_diff}
Dingkun Zhang, Sijia Li, Chen Chen, Qingsong Xie, and Haonan Lu.
\newblock Laptop-diff: Layer pruning and normalized distillation for compressing diffusion models.
\newblock \emph{arXiv preprint arXiv:2404.11098}, 2024.

\bibitem[Zhang et~al.(2023)Zhang, Herrmann, Hur, Cabrera, Jampani, Sun, and Yang]{zhang2023tale}
Junyi Zhang, Charles Herrmann, Junhwa Hur, Luisa~Polania Cabrera, Varun Jampani, Deqing Sun, and Ming-Hsuan Yang.
\newblock A tale of two features: Stable diffusion complements dino for zero-shot semantic correspondence.
\newblock \emph{NeurIPS}, 2023.

\bibitem[Zhang and Agrawala(2023)]{zhang2023adding}
Lvmin Zhang and Maneesh Agrawala.
\newblock Adding conditional control to text-to-image diffusion models, 2023.

\bibitem[Zheng et~al.(2023)Zheng, Nie, Vahdat, Azizzadenesheli, and Anandkumar]{zheng2023_fastsampling}
Hongkai Zheng, Weilie Nie, Arash Vahdat, Kamyar Azizzadenesheli, and Anima Anandkumar.
\newblock Fast sampling of diffusion models via operator learning.
\newblock In \emph{Proceedings of the 40th International Conference on Machine Learning}. JMLR.org, 2023.

\bibitem[Zheng et~al.(2024)Zheng, Hu, Fan, Wang, Ding, Tao, and Cham]{zheng2024TCD}
Jianbin Zheng, Minghui Hu, Zhongyi Fan, Chaoyue Wang, Changxing Ding, Dacheng Tao, and Tat-Jen Cham.
\newblock Trajectory consistency distillation.
\newblock \emph{arXiv preprint arXiv:2402.19159}, 2024.

\end{thebibliography}
}

\clearpage
\setcounter{page}{1}
\maketitlesupplementary

\setcounter{table}{0}
\setcounter{figure}{0}
\renewcommand{\thefigure}{S\arabic{figure}}
\renewcommand{\thetable}{S\arabic{table}}

\appendix

We provide implementation details (see \cref{imp_details}) and additional results (see \cref{app_additional_results}) for our \ourmethod image generation method with \textbf{loop-free} inference. Subsequently, we discuss the limitations and future work (see~\cref{sec:limitation}), broader impacts (see~\cref{sec:impacts}), ethical statement (see~\cref{sec:ethical}), and reproducibility statement (see~\cref{sec:statement}).

\section{Implementation Details}
\label{imp_details}

\subsection{Evaluation Datasets and Metrics}

\minisection{Datasets.}
We conduct comparisons on four datasets to evaluate the density of image generation: AFHQ~\citep{choi2020stargan}, CelebA-HQ~\citep{karras2017progressive}, DrawBench~\citep{saharia2022photorealistic}, and PartiPrompts~\citep{yu2022scaling}. Since the AFHQ and CelebA-HQ datasets contain animals and human faces respectively, we utilize text prompts with the format: "a photo of $<$cat/dog/wild animal$>$" and "a photo of $<$man/woman$>$".

\minisection{Metrics.}
We leverage code from the popular GitHub repository ``StudioGAN''~\citep{kang2023studiogan, Kang2021RebootingAA, Kang2020ContraGANCL}
\footnote{\url{https://github.com/POSTECH-CVLab/PyTorch-StudioGAN}}
to calculate three metrics:  Precision Recall~\cite{kynkaanniemi2019improved}, Density, and Coverage~\citep{naeem2020reliable}. 
For FID~\citep{heusel2017gans} and Clipscore~\citep{hessel2021clipscore} metrics, we employ the official envaluation code from GigaGAN~\cite{kang2023scaling}
\footnote{\url{https://github.com/lucidrains/gigagan-pytorch}}

\subsection{Baseline Implementations}
We use the official implementation of Instaflow~\citep{liu2023instaflow}
\footnote{\url{https://github.com/gnobitab/InstaFlow}}
, 
LCM~\citep{luo2023latent}
\footnote{\url{https://latent-consistency-models.github.io/}}
, SD-Turbo~\citep{sauer2023adversarial}
\footnote{\url{https://github.com/Stability-AI/generative-models}}
, and SwiftBrush~\cite{nguyen2023swiftbrush}\footnote{\url{https://github.com/VinAIResearch/SwiftBrush}}.
For SwiftBrushv2~\cite{dao2024swiftbrushv2}, we re-implemented the work using the same amount of training data and computational resources as our method. 
All experiments are conducted at a standard resolution of 512×512 pixels on a single 3090 GPU device.

\subsection{Training Details}
We use Stable Diffusion 2.1 (SD 2.1)\footnote{\url{https://huggingface.co/stabilityai/stable-diffusion-2-1-base}}  to initialize  the teacher SD generator and SD-LoRA generator, and  the student generator. 
We implement our method with PyTorch, and use the Adam optimizer~\citep{kingma2017adam} with $\beta_1=0.9$ and $\beta_2=0.999$ to train both the student generator and SD-LoRA generator. When calculating the VSD loss $\mathcal{L}_{vsd}$, we use classifier-free guidance with a value of 4.5 like SwiftBrush~\cite{nguyen2023swiftbrush} for both the teacher SD generator and the SD-LoRA generator.

\textbf{JourneyDB Datasets.} In JourneyDB datasets~\citep{pan2023journeydb}, there are 4M (4,189,737) captions in the training sets. We remove duplicate captions from the training set, leaving 1,418,723 unique captions. These captions are used as prompts to train the student generator. We train our model on NVIDIA 8$\times$A40 48G GPU with batch size 64 and take 3 epochs.

\subsection{The Green and Red Arrows in~\cref{pipline_baselines} and~\cref{pipline}}
The \green{green} arrows indicate the skip connections that transfer features from the middle layers of the encoder to the corresponding decoder layers, while the \red{red} arrows represent the path where features from the final encoder layer are inputs to the decoder.

At each time step, the decoder receives both the Mid-Block outputs and skip-connection features from the encoder. Since the Mid-Block does not receive skip-connection features, it is not shown in~\cref{pipline_baselines} and~\cref{pipline} and is considered part of the encoder.

\begin{table*}[tb]
\renewcommand{\arraystretch}{.5}
\tabcolsep=0.04cm
\centering
\resizebox{0.999\linewidth}{!}{
\begin{tabular}{c|c|c|c|c|c|c|c|c|c|c|c|c|c|c|c|c|c|c}
\toprule
Dataset & {\multirow{4}{*}{\makecell{Base\\Model}}}& {\multirow{4}{*}{Step}} & {\multirow{4}{*}{Param}} & \multicolumn{5}{c|}{COCO2014-30K}  & \multicolumn{5}{c|}{\centering COCO2017-5K} & \multicolumn{2}{c|}{\makecell{Inference$\downarrow$}} & \multicolumn{2}{c|}{\makecell{Training Data}} & \multirow{4}{*}{\makecell{A100\\Days$\downarrow$}}\\
\cmidrule{1-1}\cmidrule{5-14}\cmidrule{15-18}
\diagbox{Method}{Metrics} & & & & {FID$\downarrow$} & {CLIP$\uparrow$} & {Precision$\uparrow$} & {Recall$\uparrow$} & {F1$\uparrow$}& {FID$\downarrow$} & {CLIP$\uparrow$} & {Precision$\uparrow$} & {Recall$\uparrow$} & {F1$\uparrow$}& \makecell{Time\\(ms)} & \makecell{Memory\\(GB)} & Size$\downarrow$ & \makecell{Image\\Free}\\
\midrule\midrule
SD1.5~\cite{rombach2022high} (\textit{cfg}=7.5)$^\dag$ & -- & 50 & 860M & 16.08 & 0.325 & 0.717 & 0.527 & 0.607 & 23.39 & 0.326 & 0.776 & 0.587& 0.668 & 2503.0 & 4.04 & 5B& \xmarkg & 4783\\
SD1.5~\cite{rombach2022high} (\textit{cfg}=4.5)$^\dag$ & -- & 50 & 860M & 9.90 & 0.322 & 0.727 & 0.585 & 0.648 & 19.87 & 0.323 & 0.764 & 0.649 & 0.702 & 2503.0 & 4.04 & 5B& \xmarkg & 4783\\
SD2.1~\cite{rombach2022high} (\textit{cfg}=7.5)$^\dag$ & -- & 50 & 865M & 16.10 & 0.328 & 0.723 & 0.489 & 0.583 & 25.40 & 0.328 & 0.769 & 0.561 & 0.649 & 2244.2 & 3.89 & 5B & \xmarkg & 8332\\
SD2.1~\cite{rombach2022high} (\textit{cfg}=4.5)$^\dag$ & -- & 50 & 865M & 12.22 & 0.325 & 0.734 & 0.526 & 0.614 & 22.24 & 0.298 & 0.788 & 0.606 & 0.685 & 2244.2 & 3.89 & 5B& \xmarkg & 8332\\
\midrule
{FasterD}~\cite{li2024faster} (\textit{cfg}=7.5)$^\dag$ & \multirow{2}{*}{SD1.5} & 50 & 860M & 12.93 & 0.326 & 0.693 & 0.532 & 0.601 & 23.10 & 0.325 & 0.687 & 0.601 & 0.641 & 1476.0 & 21.83 & -- & -- & --\\
{FasterD}~\cite{li2024faster} (\textit{cfg}=4.5)$^\dag$ &  & 50 & 860M & 12.05 & 0.323 & 0.672 & 0.569 & 0.617 & 22.32 & 0.322 & 0.670 & 0.638 & 0.654 & 1476.0 & 21.83 & --& -- & --\\
\cmidrule{2-2}
{FasterD}~\cite{li2024faster} (\textit{cfg}=7.5)$^\dag$ & \multirow{2}{*}{SD2.1} & 50 & 865M & 13.64 & 0.329 & 0.708 & 0.512 & 0.594 & 23.65 & 0.329 & 0.698 & 0.572 & 0.629 & 1356.0 & 21.26 & -- & -- & --\\
{FasterD}~\cite{li2024faster} (\textit{cfg}=4.5)$^\dag$ &  & 50 & 865M & 12.42 & 0.326 & 0.699 & 0.551 & 0.616 & 22.61 & 0.325 & 0.707 & 0.616 & 0.659 & 1356.0 & 21.26 & --& -- & --\\
\midrule
GigaGAN~\cite{kang2023scaling}$^*$ & GAN & 1 & 1.0B & 9.24 & 0.325 & 0.724 & 0.547 & 0.623 & -- & -- & -- & -- & -- & -- & -- & 2.7B & \xmarkg & 6250\\
\midrule\midrule
InstaFlow~\cite{liu2023instaflow}$^\dag$ & \multirow{6}{*}{SD1.5} & 1 & 0.9B & \underline{13.78} & 0.288 & 0.654 & \underline{0.521} &  \underline{0.580} & \textbf{19.00} &0.293 & 0.729 & \underline{0.613} &  \underline{0.666} & 111.3 & 3.99 & 3.2M & \xmarkg & 183.2\\
LCM~\cite{luo2023latent}$^\dag$ & & 1 & 860M & 132.09 & 0.230 & 0.109 & 0.194 & 0.140 & 143.73 & 0.229 & 0.118 & 0.291 & 0.168 &236.2 & 5.88 & 12M&\xmarkg & 1.3\\
{LCM-LoRA~\cite{luo2023lcm_lora}} $^\dag$ & & 1 & 860M & 115.21 & 0.280 & 0.069 & 0.221 & 0.105 & 126.82 & 0.280 & 0.070 & 0.265 & 0.111 & 101.4 & 4.66 & 12M&\xmarkg & 1.3 \\
{Hyper-SD~\cite{ren2024hyper}} $^\dag$ & & 1 & 860M & 20.90 & 0.325 & 0.743 & 0.324 & 0.451 & 30.45 & 0.325 & 0.799 & 0.424 & 0.554 & 117.5 & 4.54 & unk. & \xmarkg & 33.3 \\
\cmidrule{2-2}
{SD-Turbo}~\cite{sauer2023adversarial}$^\dag$ & \multirow{5}{*}{SD2.1} & 1 & 865M & 19.51 & {0.331} & 0.758 & 0.458 & 0.571 & 29.35 & 0.331 & 0.786 & 0.445 & 0.568 &140.0 & 3.86 &unk.&\xmarkg & unk.\\
{TCD}~\cite{zheng2024TCD}$^\dag$ &  & 1 & 865M & 68.01 & 0.301 & 0.234 & 0.198 & .214 & 79.15 & 0.300 & 0.298 & 0.339 & 0.317 & 103.0 & 4.43 & 5B & \xmarkg & 7.1\\
SwiftBrush~\cite{nguyen2023swiftbrush}$^\dag$ & & 1 & 865M & 17.20 & 0.301 & 0.672 & 0.458 & 0.545 & 27.18 & 0.314 & 0.729 & 0.527 & 0.612 & 95.0 & 3.85 & 1.4M & \cmarkg & 4.1\\
SwiftBrushv2~\cite{dao2024swiftbrushv2}$^\ddag$ & & 1 & 865M & 15.98 & 0.326 & \textbf{0.782} & 0.457 & 0.577 & 26.28 & 0.326 & \textbf{0.816} & 0.543 & 0.652 & 139.6 & 4.91& 1.4M& \cmarkg & 24.1\\
\cmidrule{2-2}
LCM~\cite{luo2023latent}$^\dag$ & \multirow{3.5}{*}{SDXL} & 1 & 2.6B & 73.75 & 0.285 & 0.277 & 0.254 & 0.265 & 82.74 & 0.285 & 0.344 & 0.384 & 0.363 & 661.0 & 13.84 & 12M &\xmarkg & 1.3\\
SDXL-Turbo~\cite{sauer2023adversarial}$^\dag$ &  & 1 & 2.6B & 18.98 & \textbf{0.343} & 0.765 & 0.413 & 0.536 & 29.17 & \textbf{0.343} & 0.804 & 0.518 & 0.630 & 180.7 & 9.24 & unk. & \xmarkg & unk.\\
\makecell{SDXL-Lightning}~\cite{lin2024sdxl_lightning}$^\dag$ &  & 1 & 2.57B & 20.71 & {0.331} & 0.740 & 0.388 & 0.509 & 30.75 & 0.323 & 0.760 & 0.487 & 0.594 & 181.2 &9.19 & 30M &\xmarkg & unk.\\
\midrule
LCM~\cite{luo2023latent}$^\dag$ & \multirow{5}{*}{SD1.5} & 4 & 860M & 23.21 & 0.262 & 0.666 & 0.346 & 0.455 & 40.37 & 0.303 & 0.713 & 0.460 & 0.559 & 592.3 & 5.88 & 12M &\xmarkg & 1.3\\
{LCM-LoRA~\cite{luo2023lcm_lora}} $^\dag$ & & 4 & 860M & 26.06 & 0.323 & 0.722 & 0.312 & 0.436 & 36.17 & 0.322 & 0.768 & 0.406 & 0.531 & 189.9 & 4.66 & 12M&\xmarkg & 1.3 \\
{Hyper-SD~\cite{ren2024hyper}} $^\dag$ & & 4 & 860M & 21.94 & 0.326 & 0.742 & 0.327 & 0.454 & 31.73 & 0.325 & 0.804 & 0.430 & 0.560 & 221.9 & 4.55 & unk. & \xmarkg & 33.3 \\
{PCM~\cite{wang2024phased_CM}} $^\dag$ & & 4 & 860M & 21.44 & 0.316 & 0.766 & 0.360 & 0.490 & 31.35 & 0.315 & 0.770 & 0.430 & 0.552 & 304.3 & 4.56 & 3.3M & \xmarkg & 2 \\
\cmidrule{2-2}
{SD-Turbo}~\cite{sauer2023adversarial}$^\dag$ & \multirow{2}{*}{SD2.1} & 4 & 865M & 16.14 & 0.335 & 0.633 & 0.394 & 0.468& 26,14 & 0.335 & 0.694 & 0.375 & 0.487 & 272.2 & 3.86 &unk.&\xmarkg & unk.\\
{TCD}~\cite{zheng2024TCD}$^\dag$ & & 4 & 865M & 18.06 & 0.319 & 0.761 & 0.419 & 0.540 & 27.83 & 0.318 & 0.795 & 0.507 & 0.619 & 199.2 & 4.43 & 5B &\xmarkg & 7.1\\
\cmidrule{2-2}
LCM~\cite{luo2023latent}$^\dag$ & \multirow{3.5}{*}{SDXL} & 4 & 2.6B & 17.66 & 0.327 & \underline{0.780} & 0.408 & 0.536 & 27.15 & 0.328 & 0.810 & 0.513& 0.628 &1074.3 & 13.84 & 12M &\xmarkg & 1.3\\
SDXL-Turbo~\cite{sauer2023adversarial}$^\dag$ & & 4 & 2.6B & 17.79 & \underline{0.340}  & 0.769 & 0.431 & 0.552 & 27.57 & \underline{0.341} & \underline{0.814} & 0.529 & 0.641 & 305.7 & 9.24 &unk.& \xmarkg & unk.\\
\makecell{SDXL-Lightning}~\cite{lin2024sdxl_lightning}$^\dag$ & & 4 & 2.57B & 19.82 & {0.322} & 0.715 & 0.401 & 0.514 & 29.32 & 0.333 & 0.782 & 0.457 & 0.577 & 310.9 & 9.19 &30M&\xmarkg & unk.\\
\midrule
{Ours} & SD2.1 & 1 & 865M & \textbf{13.09} & {0.313} & 0.634 & \textbf{0.622} & \textbf{0.628} & \underline{23.11}  & 0.313 & 0.697 & \textbf{0.668} & \textbf{0.682} & 164.7 & 4.98 & 1.4M & \cmarkg & 3.9 \\
\bottomrule
\end{tabular}
}
\vspace{-3mm}
\caption{Comparison of our distillation method against other works. 
 Inference Time (ms) and Memory (GB). 
$^\dag$ indicates that we report results using the provided official code and pretrained models. 
$^\ddag$ denotes that we re-implemented the work and are providing the scores.
$^*$ indicates that we report results using the provided generated images.
``unk.'' denotes unknown. 
The best and second-best scores are highlighted in \textbf{bold} and \underline{underlined}, respectively.
}
\label{tab:app_fid_clip}
\end{table*}

\subsection{Meaning of Loop-Free}
We regard iterative denoising in the vanilla multi-step DMs as a ``loop'' process, while ours does not require any iterative process. Our method denoises in parallel with 4 decoder steps, achieving ``loop-free'' generation.

We further explain~\cref{eq:our_sampling} in main paper and provide a mathematical interpretation of the 1-step inference for our 1-step encoder and 4-step decoder (i.e., \textit{K}\text{=}4) design. The student generator $\epsilon^{SG}_{\theta}$ takes as input a random noise ${\epsilon}$, also referred to as $z_K$. As shown in~\cref{pipline}, we only need to calculate  skip connections and output features of the UNet-Encoder in the initiation step (t=\textit{K}) as: $f=\epsilon^{SG\text{-}EN}_{\theta}({\epsilon},K, {y})$.
Then, the predicted noise of UNet-Decoder at step t ($t\text{=}4,3,2,1$) can be calculated as $\epsilon^t=\epsilon^{SG\text{-}DE}_{\theta}(f, t, {y})$ in parallel.
Using the DDIM scheduler, the latent at step $t$ can be written as:
\begin{equation}\label{eq:1st-step}
\resizebox{0.7\linewidth}{!}{
$z_3\text{=}\sqrt{\frac{\alpha_{3}}{\alpha_4}}{\epsilon}\text{+} \sqrt{\alpha_{3}}\left(\sqrt{\frac{1}{\alpha_{3}}\text{-}1}\text{-}\sqrt{\frac{1}{\alpha_4}\text{-}1}\right) \cdot \epsilon^{4}$},
\end{equation}
\begin{equation}\label{eq:2nd-step}
\resizebox{0.7\linewidth}{!}{
$z_2\text{=}\sqrt{\frac{\alpha_{2}}{\alpha_3}}z_3\text{+}\sqrt{\alpha_{2}}\left(\sqrt{\frac{1}{\alpha_{2}}\text{-}1}\text{-}\sqrt{\frac{1}{\alpha_3}\text{-}1}\right) \cdot \epsilon^{3}$},
\end{equation}
\begin{equation}\label{eq:3rd-step}
\resizebox{0.7\linewidth}{!}{
$z_1\text{=}\sqrt{\frac{\alpha_{1}}{\alpha_2}}z_2\text{+}\sqrt{\alpha_{1}}\left(\sqrt{\frac{1}{\alpha_{1}}\text{-}1}\text{-}\sqrt{\frac{1}{\alpha_2}\text{-}1}\right) \cdot \epsilon^{2}$},
\end{equation}
\begin{equation}\label{eq:4th-step}
\resizebox{0.7\linewidth}{!}{
$z_0\text{=}\sqrt{\frac{\alpha_{0}}{\alpha_1}}z_1 \text{+} \sqrt{\alpha_{0}}\left(\sqrt{\frac{1}{\alpha_{0}}\text{-}1}\text{-}\sqrt{\frac{1}{\alpha_1}\text{-}1}\right) \cdot \epsilon^{1}$}.
\end{equation}
With ~\cref{eq:1st-step},~\cref{eq:2nd-step},~\cref{eq:3rd-step}, and~\cref{eq:4th-step}, the inference is formulated as
\begin{equation}\label{eq:steps}
\resizebox{0.7\linewidth}{!}{
$z_0=S \epsilon + E_4 \epsilon^4 + E_3 \epsilon^3 + E_2 \epsilon^2 + E_1 \epsilon^1$}.
\end{equation}
where $\scriptstyle S=\sqrt{\frac{\alpha_0}{\alpha_4}}$, $\scriptstyle E_1=\sqrt{\alpha_0}\left(\sqrt{\frac{1}{\alpha_0}-1}-\sqrt{\frac{1}{\alpha_1}-1}\right)$,
and $\scriptstyle E_t=\sqrt{\frac{\alpha_0}{\alpha_{t}}}\left(\sqrt{\frac{1}{\alpha_{t-1}}-1}-\sqrt{\frac{1}{\alpha_t}-1}\right), t\in[2,4]$.
Since $\epsilon^t$ can be computed in parallel, we achieve a one-step inference. 

This design reduces inference time with one-step sampling, incurs only a minimal increase in memory usage, and achieves significantly improved generation quality.

\subsection{\cref{tab:dns_cvg} Explanation}
Since DrawBench and PartiPrompts are prompt datasets, we used the SD2.1 samples as the ``GroundTruth''. The FID, Density, and Coverage metrics computed for the SD2.1 model, when calculated with themselves, are 0 and 1, respectively.

\begin{table*}[tb]
\renewcommand{\arraystretch}{0.8}
\tabcolsep=0.01cm
\centering
\resizebox{1.0\linewidth}{!}{
\begin{tabular}{c|c|c|c|c|c|c|c|c|c|c|c|c|c|c|c}
\toprule
Dataset & \multirow{4}{*}{\makecell{Base\\Model}} & \multirow{4}{*}{Step} & \multicolumn{3}{c|}{AFHQ} & \multicolumn{3}{c|}{CelebA-HQ} & \multicolumn{3}{c|}{DrawBench}& \multicolumn{3}{c|}{PartiPrompts}& {\makecell{Training\\Data}} \\
\cmidrule{1-1} \cmidrule{4-16}
\diagbox{Method}{Metrics} & &	& FID$\downarrow$ & Density$\uparrow$ & Coverage$\uparrow$ & FID$\downarrow$ & Density$\uparrow$ & Coverage$\uparrow$& FID$\downarrow$ & Density$\uparrow$ & Coverage$\uparrow$& FID$\downarrow$ & Density$\uparrow$ & Coverage$\uparrow$ & \makecell{Image\\Free}\\
\midrule\midrule
SD1.5~\cite{rombach2022high} (\textit{cfg}=7.5)$^\dag$ & -- & 50 & 47.16 & 0.066 & 0.030 & 93.94 & 0.053 & 0.013 & 11.95 & 0.510 & 0.622 & 7.36 & 0.730 & 0.887  & \xmarkg \\
SD2.1~\cite{rombach2022high} (\textit{cfg}=7.5)$^\dag$ & -- & 50 & 51.67 & 0.053 & 0.022 & 89.57 & 0.018 & 0.013 & 0 & 1 & 1 & 0 & 1 & 1  & \xmarkg \\
\midrule\midrule
InstaFlow~\cite{liu2023instaflow}$^\dag$ & \multirow{5}{*}{SD1.5} & 1 & \textbf{51.97} & 0.058 & \underline{0.029} & 131.99 & 0.026 & 0.007 & 25.08 & 0.223 & 0.337 & 17.64 & 0.457 & 0.670  & \xmarkg \\
LCM~\cite{luo2023latent}$^\dag$ &  & 1  & 155.63 & 0.012 & 0.033 & 165.74 & 0.001 & 0.004 & 120.98 & 0.058 & 0.014 & 95.65 & 0.095 & 0.072 & \xmarkg\\
{LCM-LoRA~\cite{luo2023latent}$^\dag$} &  & 1  & 144.15 & 0.002 & 0.001 & 249.82 & 0.009 & 0.001 & 115.63 & 0.019 & 0.014 & 92.52 & 0.057 & 0.077 & \xmarkg\\
{Hyper-SD~\cite{ren2024hyper}$^\dag$} &  & 1  & 72.07 & 0.064 & 0.018 & 126.65 & 0.055 & 0.013 & 26.04 & 0.397 & 0.385 & 17.50 & 0.677 & 0.530 & \xmarkg\\
\cmidrule{2-2}
SD-Turbo~\cite{sauer2023adversarial}$^\dag$ & \multirow{3.5}{*}{SD2.1} & 1 & 77.75 & \textbf{0.142} & 0.033 & 146.22 & 0.047 & 0.006 & 25.75 & 0.597 & 0.488 & 17.40 & 0.770 & 0.775 & \xmarkg\\
{TCD}~\cite{zheng2024TCD}$^\dag$ &  & 1 & 96.00 & 0.021 & 0.009 & 129.86 & 0.018 & 0.002 & 69.46 & 0.067 & 0.081 & 53.35 & 0.184 & 0.264 & \xmarkg\\
SwiftBrush~\cite{nguyen2023swiftbrush}$^\dag$ &  & 1 & 67.60 & 0.039 & 0.014 &  144.03  & 0.014 & 0.002 & 21.48 & 0.402 & 0.441 & {14.43} & 0.579 & 0.737 & \cmarkg \\
SwiftBrushv2~\cite{dao2024swiftbrushv2}$^\ddag$ &  & 1 & 64.99 & 0.110 & 0.025 & 131.89 & 0.055 & 0.012 & {18.57} & \underline{0.682} & \underline{0.597} & {11.32} & {0.850} & \textbf{0.865}  & \cmarkg \\
\cmidrule{2-2}
LCM~\cite{luo2023latent}$^\dag$ & \multirow{3.5}{*}{SDXL} & 1 & 106.39 & 0.022 & 0.010 & 211.19 & 0.004 & 0.001 & 83.97 & 0.043 & 0.055 & 64.31 & 0.130 & 0.226  & \xmarkg \\
SDXL-Turbo~\cite{sauer2023adversarial}$^\dag$ &  & 1 & 77.88 & 0.004 & 0.025 & 261.00 & 0.002 & 0.001 & 32.01 & 0.602 & 0.394 & 17.38 & 0.791 & 0.735 & \xmarkg\\
SDXL-Lightning~\cite{lin2024sdxl_lightning}$^\dag$ &  & 1 & 76.83 & 0.053 & 0.008 & 131.89 & {0.078} & 0.013 & 28.44 & 0.351 & 0.340 & 18.15 & 0.606 & 0.672 &  \xmarkg \\
\midrule
LCM~\cite{luo2023latent}$^\dag$ & \multirow{3.5}{*}{SD1.5} & 4  & 78.00 & 0.054 & {0.008} & 122.44 & 0.045 & \underline{0.045} & 46.23 & 0.183 & 0.187 & 26.84 & 0.512 & 0.575 & \xmarkg \\
{LCM-LoRA~\cite{luo2023latent}$^\dag$} &  & 4  & 87.54 & 0.027 & 0.005 & 228.50 & 0.045 & 0.006 & 28.45 & 0.397 & 0.362 & 19.75 & 0.732 & 0.694 & \xmarkg\\
{Hyper-SD~\cite{ren2024hyper}$^\dag$} &  & 4  & 62.67 & 0.132 & 0.031 & \textbf{76.16} & 0.096 & 0.020 & {17.53} & 0.642 & 0.569 & {11.45} & \underline{0.884} & \underline{0.853} & \xmarkg\\
{PCM~\cite{wang2024phased_CM}$^\dag$} &  & 4  & 60.02 & 0.108 & 0.026 & \underline{86.90} & \textbf{0.117} & 0.016 & \underline{17.38} & 0.673 & 0.581 & \underline{10.74} & \textbf{0.887} & \textbf{0.865} & \xmarkg\\
\cmidrule{2-2}
SD-Turbo~\cite{sauer2023adversarial}$^\dag$ & SD2.1 & 4 & 77.23 & 0.011 & 0.005 & 193.08 & 0.013 & 0.001 & 27.80 & 0.281 & 0.371 & 22.84 & 0.500 & 0.648 & \xmarkg \\
{TCD}~\cite{zheng2024TCD}$^\dag$ &  & 4 & 54.64 & 0.063 & 0.024 & 103.36 & 0.078 & 0.014 & \textbf{11.46} & 0.623 & 0.431 & \textbf{7.90} & 0.882 & 0.922 & \xmarkg\\
\cmidrule{2-2}
LCM~\cite{luo2023latent}$^\dag$ & \multirow{3.5}{*}{SDXL} & 4 & 78.72 & \underline{0.136} & 0.030 & {120.19} & 0.060 & 0.012 & 24.97 & 0.431 & 0.443 & 16.27 & 0.682 & 0.759  & \xmarkg \\
SDXL-Turbo~\cite{sauer2023adversarial}$^\dag$ &  & 4  & 79.00 & 0.067 & 0.027 & 192.97 & 0.013 & 0.010 & 30.05 & 0.466 & 0.373 & 17.12 & 0.741 & 0.742 & \xmarkg \\
SDXL-Lightning~\cite{lin2024sdxl_lightning}$^\dag$ &  & 4 & 80.23 & 0.002 & 0.001 & 130.76 & 0.035 & 0.006 & 44.52 & 0.107 & 0.132 & 36.12 & 0.248 & 0.395 & \xmarkg \\
\midrule
Ours & SD2.1 & 1 & \underline{54.48} & {0.068} & \textbf{0.071} & {116.82} & \underline{0.116} & \textbf{0.068} & {21.10} &\textbf{0.685} & \textbf{0.616} & {16.28} & {0.852} & {0.840} & \cmarkg \\
\bottomrule
\end{tabular}
} \vspace{-2mm}
\caption{\small Quantitative comparison of our distillation method with other approaches based on FID, Density, and Coverage metrics to assess diversity. $^\dag$ indicates that we report results using the provided official code and pretrained models. 
$^\ddag$ denotes that we re-implemented the work and are providing the scores. The best and second-best numbers are marked with \textbf{bold} and \underline{underlined}, respectively.}
\label{tab:app_dns_cvg}
\end{table*}

\section{Additional Results}
\label{app_additional_results}

\subsection{Our Additional Samples}
\begin{figure*}[ht]
\begin{center}
\centerline{\includegraphics[width=0.96\linewidth]{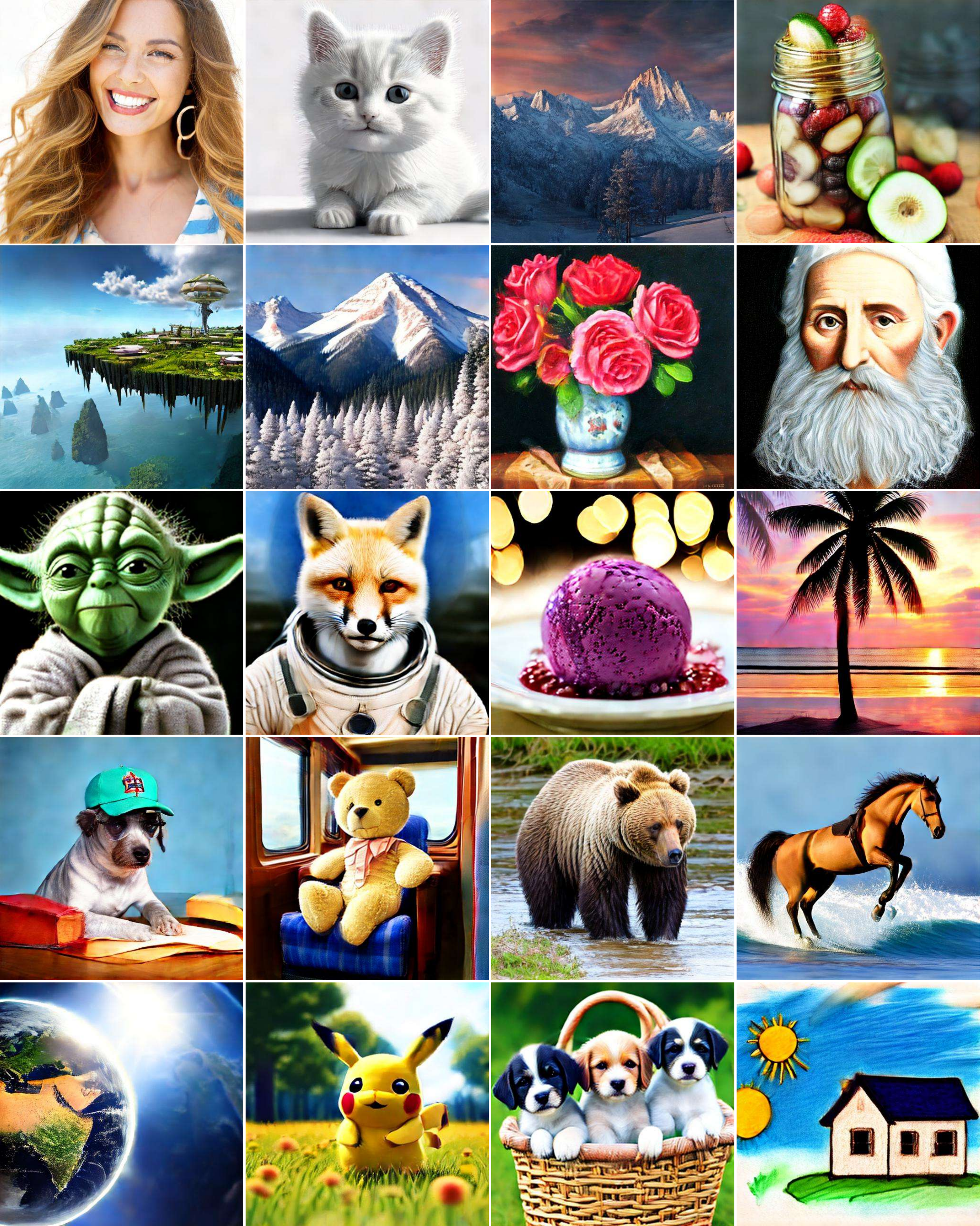}}
\caption{Our additional samples.}
\label{add_samples}
\end{center}
\end{figure*}

\begin{figure*}[ht]
\begin{center}
\centerline{\includegraphics[width=0.96\linewidth]{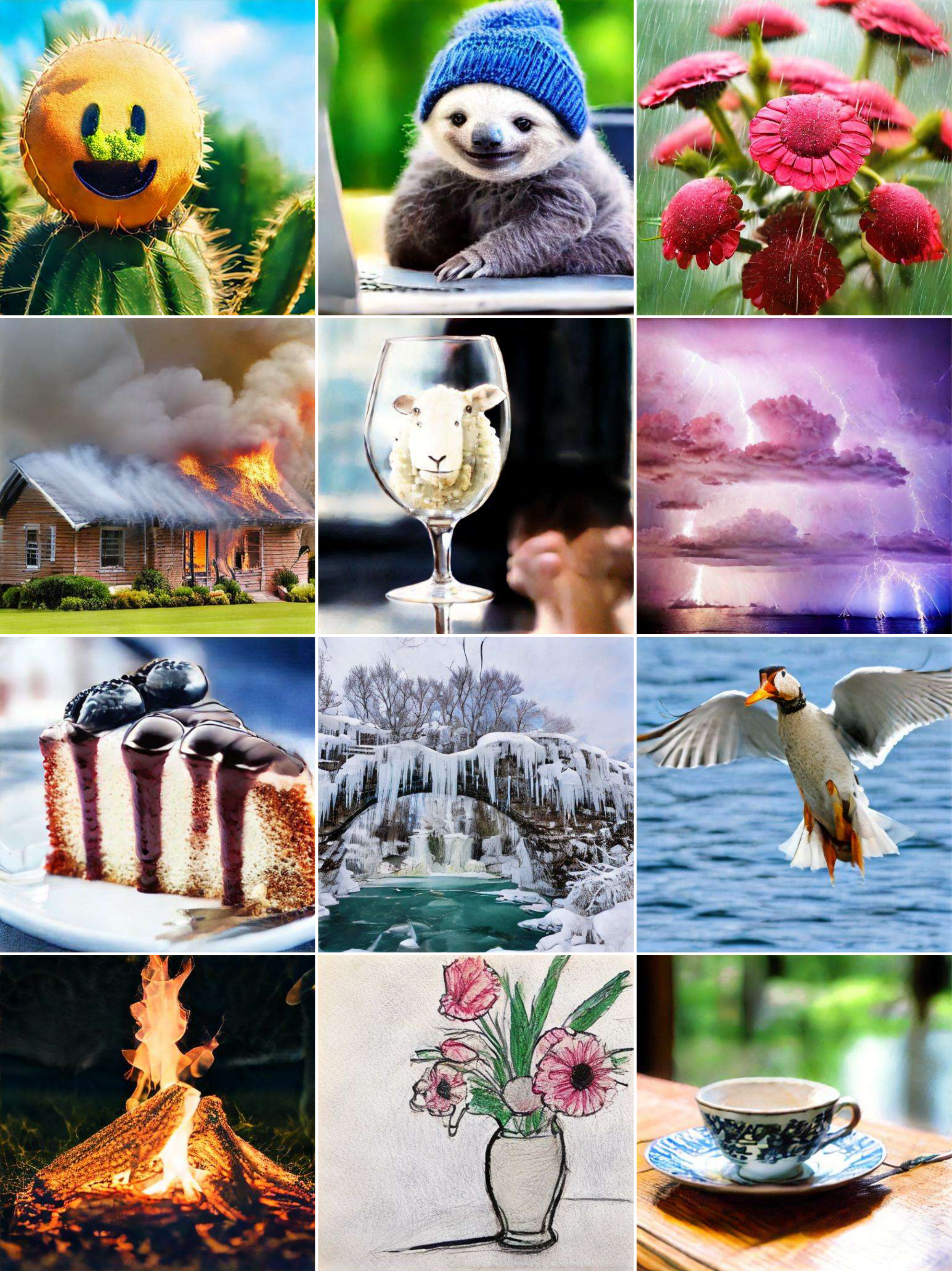}}
\caption{Our additional samples.}
\label{add_samples2}
\end{center}
\end{figure*}

\cref{add_samples,add_samples2} show our additional samples conditioned on 32 random text prompts.

\subsection{Diversity Comparison}
\label{app_diversity_comparison}
In \cref{one_step_results1} and \cref{one_step_results2}, we show additional results for the qualitative comparison of diversity. We observe that both LCM and SD-Turbo tend to generate images with limited diversity, while they generate high-quality results. SwiftBrushv2 is initialized by SD-Turbo, and thus inherits its diversity problem. In contrast, we are able to produce more realistic and diverse results, and close to the ones of SD, indicating our advantage over the baselines.

\begin{figure*}[ht]
\centering
\centerline{\includegraphics[width=0.72\linewidth]{fig_appendix/diversity_cvpr2025_compressed.pdf}}
\caption{Diversity comparison. Our results are close to the one of SD.}
\label{one_step_results1}
\end{figure*}

\begin{figure*}[ht]
\centering
\centerline{\includegraphics[width=0.72\linewidth]{fig_appendix/diversity2_cvpr2025_compressed.pdf}}
\caption{Diversity comparison. Our results are close to the one of SD.}
\label{one_step_results2}
\end{figure*}

\begin{figure*}[ht]
\centering
\centerline{\includegraphics[width=0.72\linewidth]{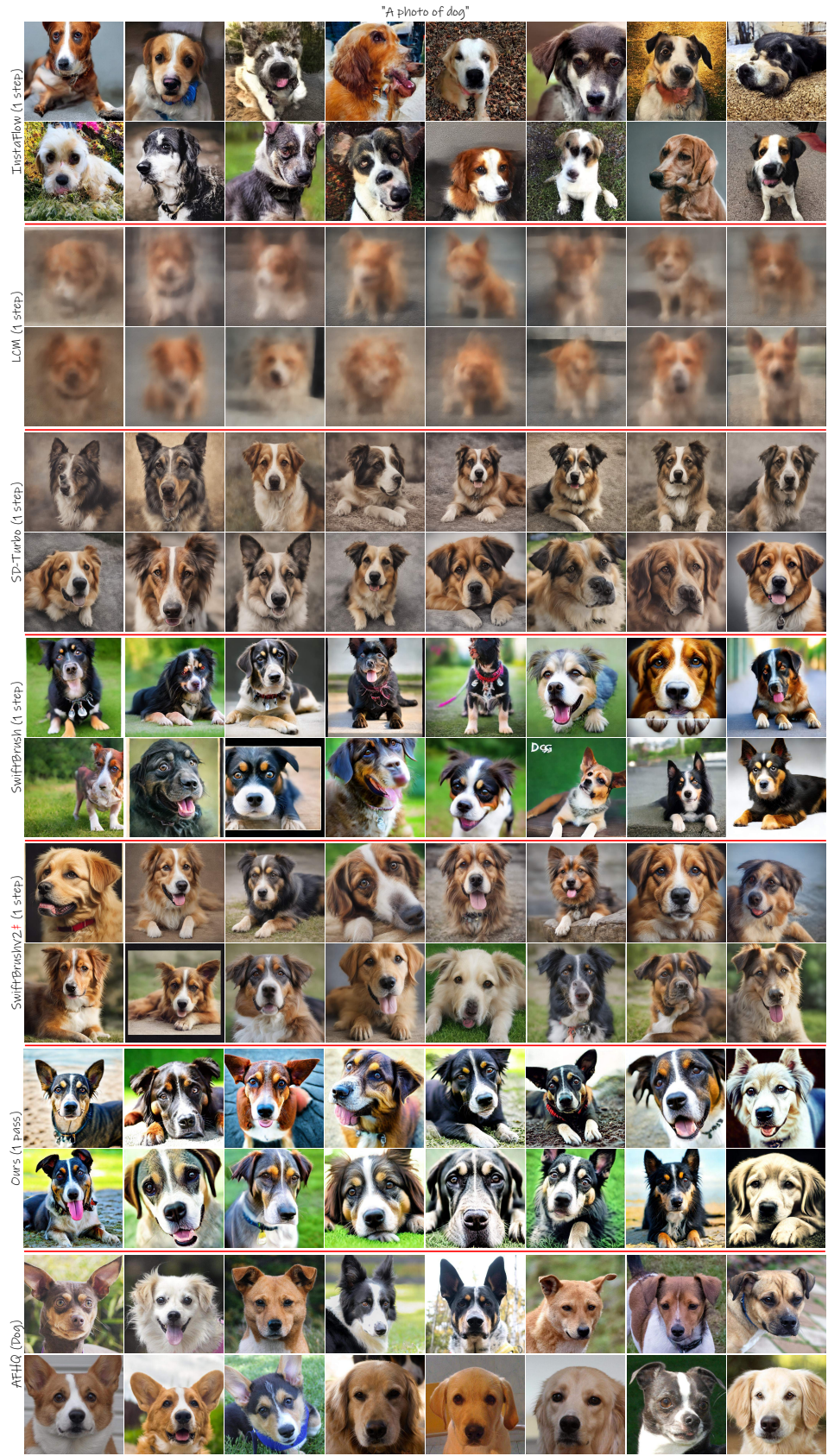}}
\caption{Diversity comparison. Our results are close to the ones of AFHQ.}
\label{afhq}
\end{figure*}

For the quantitative comparison of diversity in Tab.~\ref{tab:dns_cvg} of the main paper, as shown in \cref{afhq}, our results are closer to the real images from AFHQ.

\subsection{Iteration Qualitative Results}
\begin{figure*}[ht]
\begin{center}
\centerline{\includegraphics[width=0.96\linewidth]{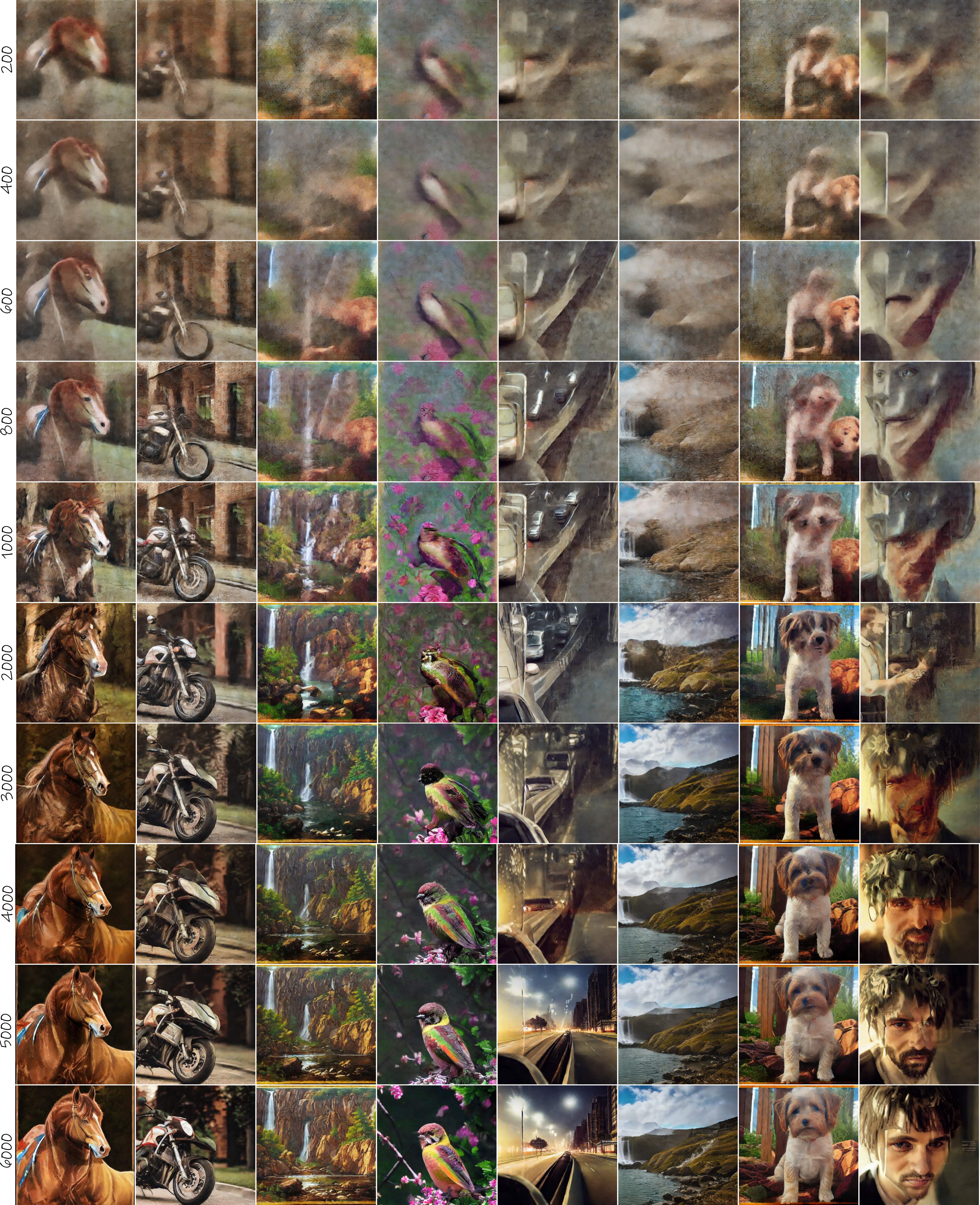}}
\caption{Iteration qualitative results.}
\label{iteration2}
\end{center}
\end{figure*}

For a better demonstration of the iterative process, we present qualitative results at early 6000 steps in \cref{iteration2}.
At 1000 iterations, the model already learned meaningful texture information (see \cref{iteration2} (the fifth row)).

\subsection{Comparison with Additional SD-based Models}
To demonstrate the effectiveness of our distillation method, we compare it with FasterDiffusion (FasterD)~\cite{li2024faster}, which shares encoder features at certain adjacent time steps and performs the decoder in parallel at these steps.
FasterD~\cite{li2024faster} accelerates 50-step sampling by 1.8x during inference while maintaining generation quality (FID = 12.42 and 22.61 for COCO2014 and COCO2017) (see~\cref{tab:app_fid_clip}). We further include the comparison with LCM-LoRA~\cite{luo2023lcm_lora}\footnote{\url{https://huggingface.co/latent-consistency/lcm-lora-sdv1-5}}, PCM~\cite{wang2024phased_CM}\footnote{\url{https://github.com/G-U-N/Phased-Consistency-Model}}, Hyper-SD~\cite{ren2024hyper}\footnote{\url{https://huggingface.co/ByteDance/Hyper-SD}}, and TCD~\cite{zheng2024TCD}\footnote{\url{https://github.com/jabir-zheng/TCD}}. 

Our method achieves one-step inference while preserving quality (FID = 13.09 and 23.11 for COCO2014 and COCO2017) and outperforms LCM-LoRA, PCM, Hyper-SD and TCD across FID, Recall, and F1 evaluation metrics (see~\cref{tab:app_fid_clip}), demonstrating its advantage over traditional acceleration methods.

\subsection{Comparison with SDXL-based Models}
We compare our distillation method against other works based on SD, similar to state-of-the-art methods~\cite{nguyen2023swiftbrush,dao2024swiftbrushv2}, in the main paper.

\begin{figure*}[ht]
\centering
\centerline{\includegraphics[width=0.72\linewidth]{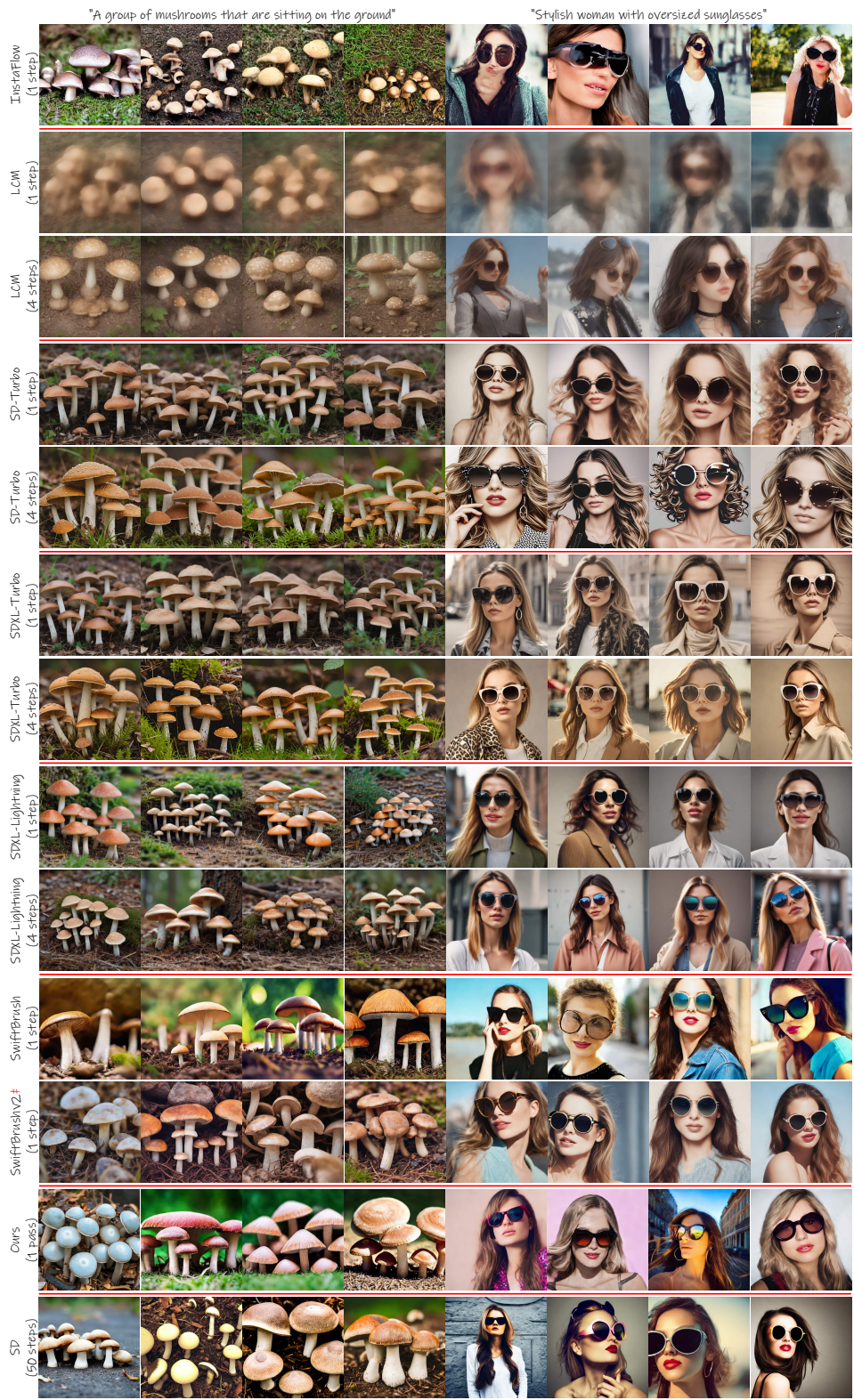}}
\caption{Qualitative and diversity comparison. Our results are close to the one of SD.}
\label{fig:sdxl}
\end{figure*}

To make a full comparison, we further include SDXL-based distillation methods, such as LCM (SDXL)~\cite{luo2023latent}\footnote{\url{https://huggingface.co/latent-consistency/lcm-sdxl}}, SDXL-Turbo~\cite{sauer2023adversarial}\footnote{\url{https://huggingface.co/stabilityai/sdxl-turbo}}, and SDXL-Lightning~\cite{lin2024sdxl_lightning}\footnote{\url{https://huggingface.co/ByteDance/SDXL-Lightning}}. 
Note that the parameter scale of these models exceeds 2.5 billion, comparing with conventional SD-based models which are often below 1 billion. 
Qualitative and quantitative results are demonstrated in \cref{tab:app_fid_clip,tab:app_dns_cvg}, and~\cref{fig:sdxl}.
As can be observe that, InstaFlow~\cite{liu2023instaflow}, LCM~\cite{luo2023latent} (1-step), and SwiftBrush~\cite{nguyen2023swiftbrush} face challenges in generating high-quality images. 
LCM~\cite{luo2023latent} (4-step), SD-Turbo~\cite{sauer2023adversarial}, SDXL-Turbo~\cite{sauer2023adversarial}, SDXL-Lightning~\cite{lin2024sdxl_lightning}, and SwiftBrushv2~\cite{dao2024swiftbrushv2} tend to generate results with similar scenes and identities giving the same prompt, leading to the lack of generation diversity. In contrast, our results are closer to the generation quality and diversity of original SD model.
Note that, due to the large amount of parameters of the SDXL model, distillting them into 1-step models~\cite{lin2024sdxl_lightning} generally requires 8$\times$80GB GPUs with batch size as 8. We aim to develop more efficient distillation approaches in the future for extremely large T2I diffusion models to reduce the time and space complexity.

\section{Limitations, and Future Work} 
\label{sec:limitation}

The present study focuses on implementing a loop-free inference with a shared encoder strategy exclusively in image-free distillation. However, we posit that adopting this shared encoder strategy in image-dependent distillation could yield loop-free sampling, thereby enhancing inference speed without compromising on generation quality. This primarily requires engineering efforts.

\section{Broader Impacts}
\label{sec:impacts}
\textit{\ourmethod} designs a loop-free inference strategy in text-to-image synthesis to improve sampling speed. However, it also carries potential negative implications. It could be used to generate false or misleading images, thereby spreading misinformation. If \textit{\ourmethod} is applied to generate images of public figures, it poses a risk of infringing on personal privacy. Additionally, the automatically generated images may also touch upon copyright and intellectual property issues.

\section{Ethical Statement}
\label{sec:ethical}
We acknowledge the potential ethical implications of deploying generative models, including issues related to privacy, data misuse, and the propagation of biases. All models used in this paper are publicly available, as well as the base training scripts. We will release the modified codes to reproduce the results of this paper. We also want to point out the potential role of customization approaches in the generation of fake news, and we encourage and support responsible usage of these models. Finally, we think that awareness of open-world forgetting can contribute to safer models in the future, since it encourages a more thorough investigation into the unpredictable changes occurring when adapting models to new data. 

\section{Reproducibility Statement}
\label{sec:statement}
To facilitate reproducibility, we will make the entire source code and scripts needed to replicate all results presented in this paper available after the peer review period. We will release the code for the novel loop-free sample we have introduced. We conducted all experiments using publicly accessible datasets. Elaborate details of all experiments have been provided in the Appendices.


\end{document}